\documentclass[lettersize,journal]{IEEEtran}
\usepackage{amsmath,amsfonts}
\usepackage{array}
\usepackage[caption=false,font=normalsize,labelfont=sf,textfont=sf]{subfig}
\usepackage{textcomp}
\usepackage{stfloats}
\usepackage{url}
\usepackage{verbatim}
\usepackage{graphicx}
\usepackage{cite}
\usepackage{multirow}
\usepackage{bm}
\usepackage{booktabs}
\usepackage[colorlinks=true,linkcolor=black,hidelinks]{hyperref}
\usepackage{xcolor}
\usepackage{subfig} 
\usepackage{subfloat}
\makeatother

\usepackage[linesnumbered,ruled,vlined]{algorithm2e}
\usepackage{algpseudocode}  
\usepackage{amsmath}  

\begin{document}

\title{Socially Aware Robot Crowd Navigation via Online Uncertainty-Driven Risk Adaptation}

\author{Zhirui Sun, Xingrong Diao, Yao Wang, Bi-Ke Zhu, and Jiankun Wang, \emph{Senior Member, IEEE} 
\thanks{This work is supported by National Natural Science Foundation of China under Grant 62473191, Shenzhen Key Laboratory of Robotics Perception and Intelligence (ZDSYS20200810171800001), Shenzhen Science and Technology Program under Grant 20231115141459001, RCBS20221008093305007, Guangdong Basic and Applied Basic Research Foundation under Grant 2025A1515012998, Young Elite Scientists Sponsorship Program by CAST under Grant 2023QNRC001, and the High level of special funds (G03034K003) from Southern University of Science and Technology, Shenzhen, China. \emph{(Corresponding author: Jiankun Wang).}}
\thanks{Zhirui Sun, Xingrong Diao, Yao Wang, Bi-Ke Zhu, and Jiankun Wang are with Shenzhen Key Laboratory of Robotics Perception and Intelligence, Department of Electronic and Electrical Engineering, Southern University of Science and Technology, Shenzhen, China (e-mail: \url{wangjk@sustech.edu.cn}).}%
\thanks{Zhirui Sun and Jiankun Wang are also with Jiaxing Research Institute, Southern University of Science and Technology, Jiaxing, China.}%
}

\maketitle

\begin{abstract} 
Navigation in human-robot shared crowded environments remains challenging, as robots are expected to move efficiently while respecting human motion conventions. However, many existing approaches emphasize safety or efficiency while overlooking social awareness. This article proposes Learning-Risk Model Predictive Control (LR-MPC), a data-driven navigation algorithm that balances efficiency, safety, and social awareness. LR-MPC consists of two phases: an offline risk learning phase, where a Probabilistic Ensemble Neural Network (PENN) is trained using risk data from a heuristic MPC-based baseline (HR-MPC), and an online adaptive inference phase, where local waypoints are sampled and globally guided by a Multi-RRT planner. Each candidate waypoint is evaluated for risk by PENN, and predictions are filtered using epistemic and aleatoric uncertainty to ensure robust decision-making. The safest waypoint is selected as the MPC input for real-time navigation. Extensive experiments demonstrate that LR-MPC outperforms baseline methods in success rate and social awareness, enabling robots to navigate complex crowds with high adaptability and low disruption. A website about this work is available at \href{https://sites.google.com/view/lr-mpc}{https://sites.google.com/view/lr-mpc}.
\end{abstract}

\begin{IEEEkeywords}
Risk assessment, uncertainty filtering, socially aware, crowd navigation, model predictive control.
\end{IEEEkeywords}

\section{Introduction}
\begin{figure}[htb]
\centering
    \includegraphics[width=0.7\columnwidth]{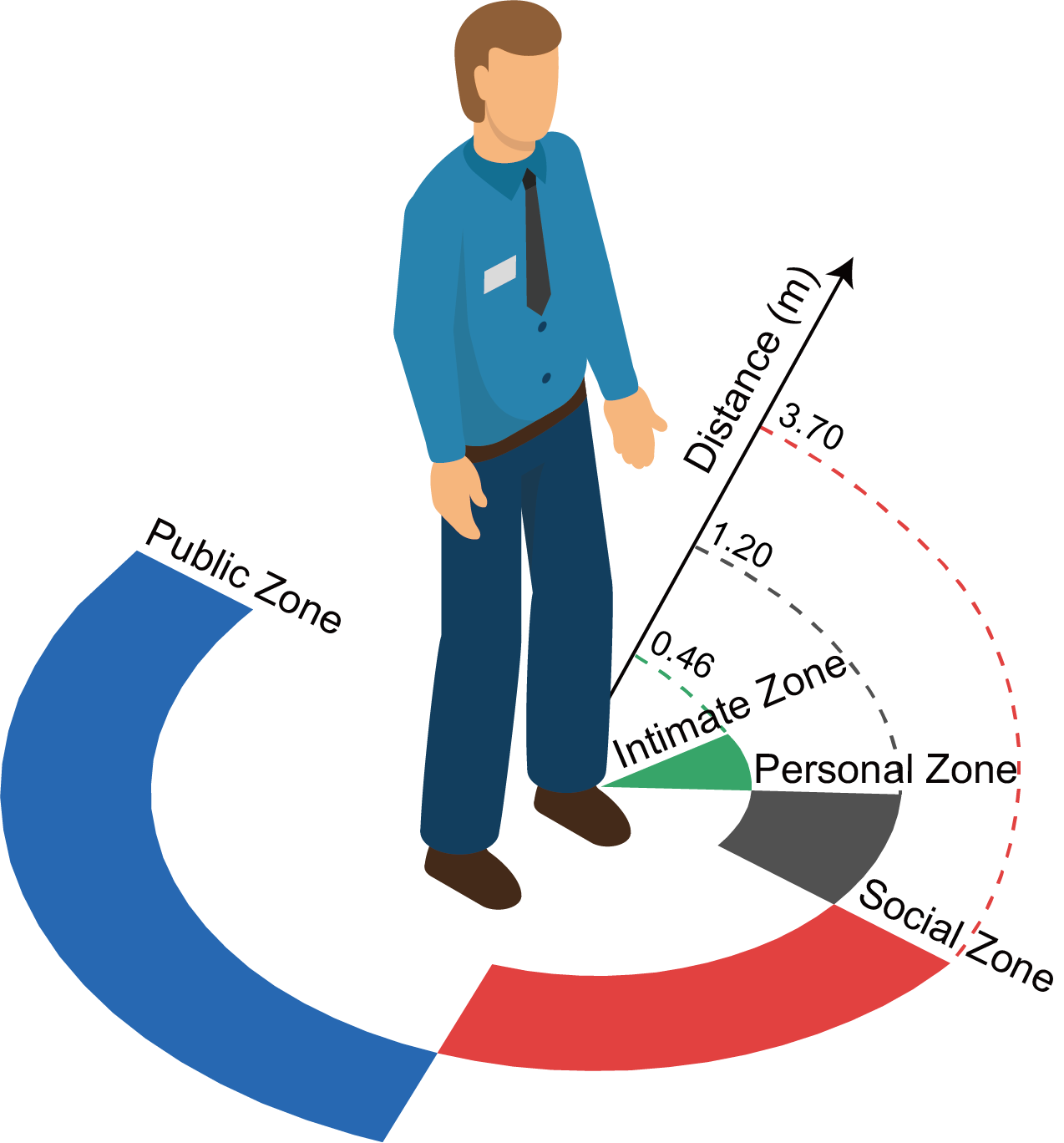}
\caption{Personal Space Zones in Human Proxemics.}
\label{social-distance}
\end{figure}
\IEEEPARstart{W}{ith} the rapid advancement of service robotics \cite{TMC1, TMC2}, the focus has gradually shifted from industrial to social applications, where robots are expected to coexist and interact closely with humans. Unlike structured industrial environments, social settings demand that mobile robots navigate crowded and dynamic spaces in a manner that minimizes disruption and preserves human comfort. In such contexts, achieving safe and socially aware navigation among moving agents remains a critical yet challenging problem.

To address the challenge of navigating in dense crowds, various approaches have been proposed. Traditional reaction-based methods such as Optimal Reciprocal Collision Avoidance (ORCA) \cite{orca} and Social Force (SF) \cite{sf} rely on immediate interaction rules to compute collision-free motions, while learning-based approaches formulate navigation as a Markov Decision Process (MDP) \cite{TMC3} and leverage neural networks for policy learning \cite{shuijing0, shuijing1}. More recent models like AttnRNN \cite{attnrnn} utilize spatiotemporal graphs to capture agent interactions over time. However, these approaches often lack explicit assessment of navigational risk, referring to the likelihood of collisions or violations of social norms such as personal space. As a result, they may produce reactive, unsafe, or socially inappropriate behaviors in human-centered environments.

Researchers in psychology have long studied human spatial needs. As illustrated in Fig. \ref{social-distance}, Hall \emph{et al}. \cite{people_distance} classified interpersonal distances into four zones: intimate, personal, social, and public. Among them, personal space denotes the area an individual perceives as their own psychological territory; intrusion into this space often triggers discomfort, anxiety, or even anger. In human-robot shared environments, humans naturally protect this space. Therefore, respecting personal space boundaries is essential for designing socially acceptable robot navigation systems. Several studies have attempted to address this by incorporating proxemic constraints into planning algorithms. Some methods introduce discomfort functions and CCR maps to model personal space violations \cite{ccr}, while others construct social interaction spaces using asymmetric Gaussian functions based on human states and orientations \cite{asymmetric_Gaussian}. Some approaches model group dynamics through Gaussian mixture-based social norms \cite{cai_gmm} or optimize kinodynamic trajectories using hybrid sampling and graph-based techniques \cite{graph_optimization}. Although these methods show improved social awareness, most rely on handcrafted social models or predefined heuristics, limiting their adaptability and generalization in crowded, dynamic environments.

To address the limitations of existing navigation methods in crowded environments, this article proposes a novel navigation algorithm that enables safe and socially aware robot behavior. The approach leverages a data-driven Probabilistic Ensemble Neural Network (PENN) \cite{penn, penn2}, trained offline using heuristic risk assessments to learn patterns of human-robot interaction. The algorithm dynamically assesses candidate waypoints at runtime, filtering out unreliable predictions based on epistemic and aleatoric uncertainties \cite{uncertainty, penn2}. Navigation decisions are guided by selecting the waypoint with the lowest risk, enabling safe, socially aware behavior and adaptation to dense crowds with minimal disruption.
The main contributions of this article are summarized as follows:
\begin{itemize}
    \item This article presents a learning-based risk-aware navigation algorithm (LR-MPC) that integrates a PENN model with MPC, enabling safe and socially aware navigation in densely crowded environments.
    \item The proposed LR-MPC algorithm combines global path guidance from Multi-RRT with real-time local risk assessment, leveraging epistemic and aleatoric uncertainty filtering to ensure robust decision-making under uncertain dynamic scenarios.
    \item LR-MPC supports adaptive online evaluation of candidate waypoints and parallel global planner updates, ensuring high responsiveness and strong generalization. Comprehensive simulation and real-world experiments validate the superiority of LR-MPC over classic and learning-based baselines in terms of success rate and social awareness.
\end{itemize}

The remainder of this article is organized as follows. Section II reviews related work on crowd navigation. Section III presents the LR-MPC algorithm, covering risk data collection, the PENN model, uncertainty-aware filtering, the Multi-RRT global planner, and the MPC controller. Section IV outlines the experimental setup and evaluates performance in both simulation and real-world scenarios. Section V provides a detailed analysis of navigation behavior and compares LR-MPC with baseline methods. Finally, Section VI concludes the article and discusses future research directions.
\section{Related Work}\label{related-work}
This section reviews existing methods for robot navigation in crowded environments, organized into two main categories: Reactive and Predictive Navigation and Learning-Based and Socially Aware Navigation. While these approaches provide valuable insights into motion planning and human-robot interaction, many still struggle to balance efficiency, safety, and social aware in dense, dynamic settings. These limitations motivate the introduction of LR-MPC, a risk-adaptive navigation algorithm that integrates risk assessment with uncertainty-aware decision-making.
\subsection{Reactive and Predictive Navigation}
Traditional navigation in crowded environments primarily relies on reactive and predictive planning algorithms. Reactive methods generate control actions directly based on the robot’s current state and sensor inputs without explicitly reasoning about the future. For instance, the Dynamic Window Approach (DWA) \cite{dwa} evaluates a set of velocity commands to select the one that ensures collision avoidance while guiding the robot toward the goal. Similarly, velocity-based models such as Reciprocal Velocity Obstacles (RVO) \cite{rvo} and ORCA \cite{orca} represent nearby agents as dynamic constraints and assume reciprocal responsibility for collision avoidance. These approaches are computationally efficient and widely adopted in real-time systems. However, their effectiveness is often limited by strong assumptions, such as reciprocal compliance from humans. These assumptions may break down in densely crowded or socially unaware settings. Moreover, SF \cite{sf} models interpret human-robot interactions through artificial forces but tend to be overly sensitive to parameter tuning and often cause suboptimal, overly conservative paths in crowded environments.

Predictive methods, in contrast, plan over a short temporal horizon by predicting environmental changes and optimizing the trajectory accordingly. An example is Timed Elastic Band (TEB) \cite{teb}, which models trajectory planning as a multi-objective optimization, iteratively refining time-parameterized poses to produce smooth, collision-free, and kinodynamically feasible paths. Building on this idea, Timed-ESDT \cite{Timed-ESDT} improves planning efficiency by combining smoothness priors and obstacle likelihoods in a MAP formulation, using gradient-based optimization with state-time graph search for initialization. Additionally, Multi-Risk-RRT \cite{multi-risk-rrt} introduces a multi-directional sampling strategy where independent subtrees explore the environment in parallel, incorporating predicted environmental changes to enhance planning efficiency and safety. These predictive methods improve planning efficiency in dynamic scenarios but often struggle to ensure robustness and safety when facing ambiguous and complex human behaviors.

\subsection{Learning-Based and Socially Aware Navigation}
Recent advancements have adopted learning-based methods to tackle navigation challenges in dynamic and socially complex environments. One prominent strategy involves supervised learning from expert demonstrations \cite{gmr-rrt,namr-rrt}, where expert behavior may come from rule-based planners, human teleoperation \cite{teleoperation}, or even real-world human trajectories \cite{pedestrian_trajectory1, pedestrian_trajectory2, TBD_dataset}. These methods avoid the need for online exploration, allowing policies to be trained efficiently in a data-driven manner. However, the performance of such policies is often constrained by the diversity and quality of the demonstration data. Alternatively, reinforcement learning (RL) methods leverage interaction within simulated environments, balancing exploration and exploitation to optimize navigation policies. RL can develop navigation behaviors that outperform manually designed or imitation-based approaches by sampling various scenarios.  
For instance, decentralized structural-Recurrent Neural Networks (DS-RNN) \cite{shuijing1} have been employed to model spatiotemporal relationships among agents within model-free deep RL algorithms, enabling robots to navigate effectively in dense and partially observable crowds without explicit expert supervision. Other approaches embed velocity-obstacle-inspired reward functions into the learning process \cite{drl-vo}, resulting in control policies that generalize across varying crowd densities and previously unseen environments. While these methods improve safety and adaptability, they typically lack explicit mechanisms for evaluating risk and uncertainty, often struggling to maintain socially aware behaviors in ambiguous or unpredictable human interactions.

Beyond navigation safety and adaptability, an emerging focus lies in socially aware robot navigation \cite{social_principles, socially_aware}, where robots are expected to respect human personal space \cite{ccr}, maintain appropriate distances, and adhere to implicit social norms. To this end, many algorithms incorporate spatial representations inspired by proxemics, human motion cues, or discomfort models into the learning process. 
Neural attention mechanisms and spatiotemporal interaction graphs \cite{attnrnn} are also widely adopted to capture salient social cues and guide decision-making.
SACSoN \cite{Sacson} introduces a vision-based navigation policy trained to minimize counterfactual perturbation, enabling socially unobtrusive behavior through data-driven modeling of human-robot interactions.
SocialGAIL \cite{SocialGAIL} leverages Generative Adversarial Imitation Learning with attention-based graph networks to generate realistic human behaviors, enhancing the quality of navigation policy training. DR-MPC \cite{DR-MPC} combines model predictive control with deep reinforcement learning, using real-world data and safety-aware mechanisms to improve policy efficiency and robustness. 
Despite these advancements, existing socially aware navigation approaches still face limitations, notably in reliably handling diverse real-world social behaviors, adapting to complex crowd environments, and explicitly assessing risk and uncertainties during navigation.

Motivated by the need for safer and more socially aware robot navigation in dense crowd environments, this article proposes a novel navigation algorithm, LR-MPC. By integrating a data-driven risk prediction model with real-time motion control, LR-MPC enables the robot to quantify potential risks at candidate waypoints, assess their reliability through uncertainty-aware filtering, and select safe, socially aware waypoints in crowded environments. This comprehensive architecture combines the generalization capabilities of learning with the stability of MPC, leading to more robust and socially aware navigation.

\section{Methodology}
\begin{figure*}[htb]
\centering
    \includegraphics[width=2\columnwidth]{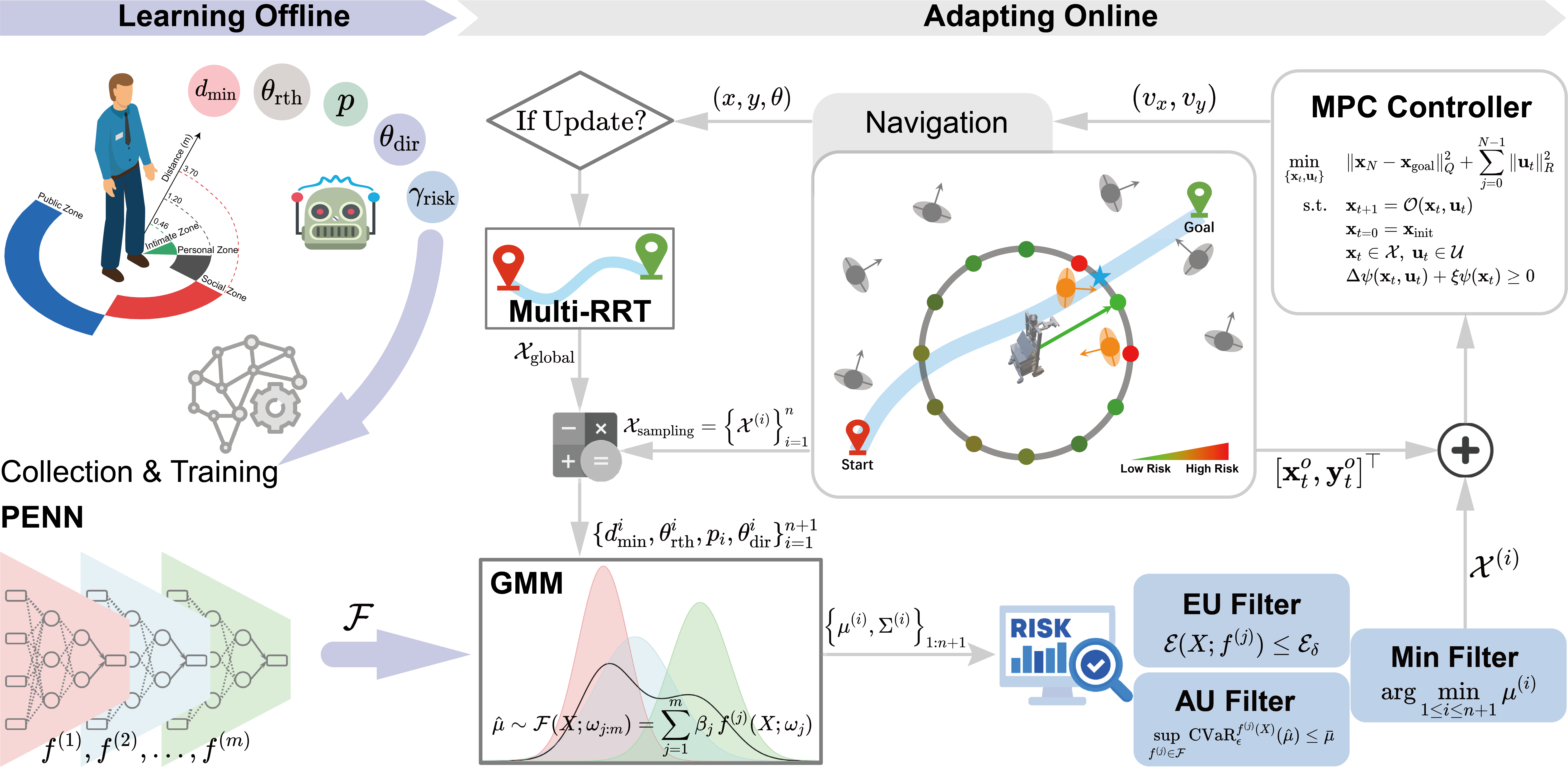}
\caption{Overview of the proposed LR-MPC algorithm. The algorithm consists of two phases: offline risk learning and online adaptive inference. In the offline phase, navigation data is collected to train a PENN model for risk assessment. During online execution, local waypoints are sampled and globally guided by the Multi-RRT planner. Each waypoint is evaluated by PENN and passed through epistemic uncertainty (EU), aleatoric uncertainty (AU), and minimum-risk filters. The selected low-risk waypoint is then sent to the MPC controller for real-time navigation. The Multi-RRT module updates dynamically based on the robot’s state and operates asynchronously to ensure efficient global guidance.}
\label{framework}
\end{figure*}

This section provides an overview of the proposed LR-MPC navigation algorithm, as shown in Fig. \ref{framework}. The algorithm includes two main phases: an offline learning phase for training the risk model and an online adaptive inference phase for real-time navigation.
In the offline phase, we construct a training dataset using the Heuristic-Risk-MPC (HR-MPC) method for risk annotation based on predefined rules (Sec. \ref{HR-MPC}). This dataset is used to train a PENN model that learns the mapping between environmental features and risk (Sec. \ref{PENN}).
In the online phase, candidate local waypoints are sampled, and a global waypoint is generated via a Multi-RRT planner (Sec. \ref{Multi-RRT}). These waypoints are evaluated for risk by the PENN model, followed by a two-stage uncertainty-aware filtering process to ensure prediction reliability (Sec. \ref{Filter}). The selected waypoint with the lowest estimated risk is then passed to the MPC controller for execution (Sec. \ref{MPC}), completing real-time navigation (Sec. \ref{Navigation}).

\subsection{HR-MPC}\label{HR-MPC}
In the HR-MPC algorithm, the waypoint risk is evaluated based on a combination of heuristic factors along the trajectory from the robot's current position to the candidate waypoint ($p_i, i \in \{1, 2, \ldots, n\}$). The total risk value $\gamma^i_{\text{risk}}$ of $p_i$ is computed by combining multiple components:

\begin{equation} \gamma^i_{\text{risk}} = \gamma^i_{p} + \gamma^i_{\text{ori}} + \gamma^i_{g} + \gamma^i_{\text{dir}}.
\end{equation}

$\gamma^i_{p}$ is the path risk, determined as the maximum inverse-distance risk along sampled waypoints between the robot and the candidate waypoint: 
\begin{equation} 
\gamma^i_{p} = \max \frac{\lambda_{\text{dist}}}{d^i_{\text{min}} + \epsilon}, 
\end{equation} 
where $d^i_{\text{min}}$ is the minimum distance from a sampled waypoint to any obstacle, $\lambda_{\text{dist}}$ is a scaling factor, and $\epsilon$ is a small constant to avoid singularity.

$\gamma^i_{\text{ori}}$ is the orientation risk, computed based on the relative angle between the robot's movement direction and the nearest human's velocity direction, penalizing head-on interactions: \begin{equation} 
\gamma^i_{\text{ori}} = \sum \exp\left(-\cos(\theta^i_{\text{rth}})\right), 
\end{equation} 
where $\theta^i_{\text{rth}}$ denotes the relative orientation at a sampled waypoint.

$\gamma^i_{g}$ is the goal distance term, which encourages selection of waypoints closer to the goal. It is defined as the Euclidean distance between the candidate waypoint $p_i$ and the goal $p_{g}$:
\begin{equation}
\gamma^i_{g} = \|p_{g} - p_i\|_2.
\end{equation}

$\gamma^i_{\text{dir}}$ is the goal direction penalty, encouraging the candidate waypoint that is aligned with the goal direction. It is computed as a penalty proportional to the misalignment (cosine similarity) between the vectors from the robot to the candidate waypoint and from the robot to the goal:
\begin{equation}
\gamma^i_{\text{dir}} = \lambda_{\text{dir}}\left(1 - \cos(\theta^i_{\text{dir}})\right),
\end{equation}
where $\theta^i_{\text{dir}}$ denotes the angle between the vector from the robot to the candidate waypoint and the vector from the robot to the goal, and $\lambda_{\text{dir}}$ is a weighting coefficient.

In the waypoint sampling process, if a sampled point collides with an obstacle, it is immediately assigned the maximum collision risk.

\subsection{Probabilistic Ensemble Neural Network Model}\label{PENN}
To model both aleatoric (data-inherent) and epistemic (model-related) uncertainties in prediction, we adopt the PENN model \cite{penn, penn2}, denoted as $\mathcal{F}$. The model predicts a risk value $\mu$, representing the continuous assessment of navigation safety at candidate waypoints.

Unlike standard deterministic neural networks, each member $f^{(j)}$ in the ensemble ($j = 1, \dots, m$), parameterized by weights $\omega_j$, predicts a full Gaussian distribution rather than a point estimate. Formally, for an input $X$, the model output is given by:
\begin{equation}
    f^{(j)}(X; \omega_j) \sim \mathcal{N}(\mu_{\omega_j}(X), \Sigma_{\omega_i}(X)),
\end{equation}
where $\mu_{\omega_j}(X)$ denotes the predicted mean and $\sigma_{\omega_j}(X)$ represents the covariance, both parameterized by the network weights $\omega_j$.

During training, each ensemble member is optimized independently using the Gaussian Negative Log-Likelihood (NLL) loss \cite{penn}, which encourages the model not only to predict accurate means but also to provide calibrated variance estimates that reflect uncertainty:

\begin{equation}
\mathcal{L}_{\text{NLL}} = \frac{1}{2} \left(
\log |\Sigma| + (\zeta - \mu)^\top \Sigma^{-1} (\zeta - \mu)
\right) + \mathcal{C},
\end{equation}
where $\zeta \in \mathbb{R}^d$ is the ground-truth, $\mu \in \mathbb{R}^d$ is the predicted mean, $\Sigma \in \mathbb{R}^{d \times d}$ is the predicted covariance matrix, and $\mathcal{C}$ is a constant independent of model parameters. For the $j$-th ensemble member, the loss is computed with $\mu := \mu_{\omega_j}(X)$ and $\Sigma := \sigma_{\omega_j}(X)$.

Once trained, the $\mathcal{F}$ ensemble members $\mathcal{F} = \{f^{(j)}, \dots, f^{(m)}\}$ collectively form a Gaussian Mixture Model (GMM) over the prediction space. The final predictive distribution $\hat{\mathcal{F}}$ is obtained by combining the predictions of all ensemble members as:
\begin{equation}
    \hat{\mu} \sim \mathcal{F}(X; \omega_{j:m}) = \sum_{j=1}^{m} \beta_j \, f^{(j)}(X; \omega_j), \quad 0 \leq \beta_j \leq 1.
\end{equation}

In this work, we adopt uniform weighting for simplicity, i.e., $\beta_j = \frac{1}{m}$ for all $j \in {1, \dots, m}$. This choice is supported by the diversity among independently trained ensemble members, as random initialization and stochastic training procedures introduce sufficient variation in their learned representations. Consequently, the disagreement across ensemble predictions naturally reflects epistemic uncertainty—capturing the model’s lack of knowledge due to limited data or insufficient model capacity.

In contrast, the tail behavior of the predictive distribution characterizes aleatoric uncertainty, which is inherent to the data and irreducible even with unlimited training samples. By incorporating both types of uncertainty within an ensemble framework, PENN provides a principled and interpretable approach for uncertainty-aware learning—particularly well-suited for risk-aware robotic navigation, where safety-critical decisions must be made under uncertainty.
\subsection{Multi-RRT}\label{Multi-RRT}
A previously proposed motion planning algorithm that combines multi-directional search with heuristic-informed sampling has demonstrated significant improvements in planning efficiency \cite{multi-rrt, multi-risk-rrt}. Building upon this foundation, the algorithm is adopted here for global path planning, with its original kinematic constraints removed to enhance flexibility. The adapted version, Multi-RRT, continuously updates the global path based on the robot’s real-time position. This update mechanism employs a parallel execution strategy inspired by existing work \cite{namr-rrt}, allowing for adaptive global planning without disrupting navigation.

\subsection{Uncertainty-Aware Filtering}\label{Filter}
To ensure the confidence and reliability of predictions made by the PENN model, we introduce a dual-level uncertainty analysis that accounts for epistemic and aleatoric uncertainties. This evaluation is crucial for validating the predicted risk values.

\subsubsection{Epistemic Uncertainty Filter}
We first quantify the epistemic uncertainty, reflecting the model’s prediction confidence. To do this, we compute the divergence among the ensemble members of PENN using the Jensen-Rényi Divergence (JRD), which is based on quadratic Rényi entropy \cite{renyi}. The closed-form expression of the JRD between Gaussian mixture components \cite{closed-form} enables efficient computation over the ensemble outputs.

The epistemic uncertainty at a given input $X$ is defined as:
\begin{equation}
    \label{eq:4}
    \mathcal{E}(X; \mathcal{F}) = -\log\left( \frac{1}{m^2} \sum_{j=1}^{m} \sum_{k=1}^{m} \mathcal{D}_{(j,k)} \right) + \frac{1}{m} \sum_{j=1}^{m} \log\left( \mathcal{D}_{(j,j)} \right),
\end{equation}
where each divergence term between two Gaussians is given by:
\begin{equation}
    \mathcal{D}_{(j,k)} := \frac{1}{|\Delta \Sigma_{(j,k)}|^{1/2}} \exp\left( -\frac{1}{2} \Delta_{(j,k)}^\top \Delta \Sigma_{(j,k)}^{-1} \Delta_{(j,k)} \right),
\end{equation}
where $\mathcal{D}{(j,k)}$ quantifies the similarity between the Gaussian outputs from the $j$-th and $k$-th ensemble members. Specifically, the mean difference is defined as $\Delta \mu{(j,k)} := \mu_{\omega_j}(X) - \mu_{\omega_k}(X)$, and the covariance combination is given by $\Delta \Sigma_{(j,k)} := \Sigma_{\omega_j}(X) + \Sigma_{\omega_k}(X)$. This divergence captures the dispersion among ensemble predictions: a higher value of $\mathcal{D}_{(j,k)}$ reflects greater disagreement, indicating lower confidence and higher epistemic uncertainty. A prediction is deemed to be out-of-distribution if the epistemic uncertainty $\mathcal{E}(X; f^{(j)})$ exceeds a pre-defined threshold $\mathcal{E}_\delta$.

\subsubsection{Aleatoric Uncertainty Filter}
Since the predictive distribution of $\hat{\mu}$ reflects both the aleatoric uncertainty from each model and the variability across ensemble members, evaluating its risk-aware behavior probabilistically is essential. For a risk tolerance level $\epsilon$, the Value at Risk (VaR) \cite{var} for an ensemble member $f^{(j)}(X)$ is defined as:
\begin{equation}
\text{VaR}^{f^{(j)}(X)}_\epsilon(\hat{\mu}) := \inf\left\{ \nu \in \mathbb{R} \,\big|\, \mathbb{P}_{f^{(j)}(X)}(\hat{\mu} > \nu) \leq \epsilon \right\}.  
\end{equation}

The VaR at level $\epsilon$ is defined as the smallest threshold $\nu$ such that the probability of the predicted variable $\hat{\mu}$ exceeding $\nu$ is no more than $\epsilon$. Intuitively, $\text{VaR}_\epsilon$ provides a probabilistic upper bound such that the predicted value exceeds this threshold with probability at most $\epsilon$.
We then define the CVaR \cite{cvar}:
\begin{equation}
\text{CVaR}^{f^{(j)}(X)}_\epsilon(\hat{\mu}) := \inf_{\varrho \in \mathbb{R}} \left[ \varrho + \frac{1}{\epsilon} \mathbb{E}_{f^{(j)}(X)}[(\hat{\mu} - \varrho)^+] \right],
\end{equation}
where $(\cdot)^+$ denotes the positive part operator, defined as $\max\{0, \cdot\}$, and $\varrho$ serves as a threshold that separates the $\epsilon$-tail of the distribution. CVaR is computed as a conditional expectation over the $\epsilon$-tail of the distribution. Therefore, the tail loss is normalized by $\epsilon$ to reflect the average outcome in the worst-case $\epsilon$-probability region. CVaR is a coherent risk measure that accounts for the average and tail-end severity of outcomes beyond the VaR.

We compute the worst-case CVaR across all ensemble members to ensure robustness under model uncertainty. If this upper bound remains below a user-defined threshold $\bar{\mu}$, it implies that with probability at least $1 - \epsilon$, the predicted risk $\hat{\mu}$ will not exceed $\bar{\mu}$ under any model in the ensemble. It guarantees distributional robustness in high-risk decision-making scenarios.
\begin{equation}
    \label{eq:8}
    \sup_{f^{(j)} \in \mathcal{F}} \text{CVaR}^{f^{(j)}(X)}_\epsilon(\hat{\mu}) \leq \sup_{f^{(j)} \in \mathcal{F}} \text{VaR}^{f^{(j)}(X)}_\epsilon(\hat{\mu}) \leq \bar{\mu},
\end{equation}
which equivalently implies:
\begin{equation}
    \inf_{f^{(j)}(X) \in \mathcal{F}} \mathbb{P}_{f^{(j)}(X)}(\hat{\mu} \leq \bar{\mu}) \geq 1 - \epsilon.
\end{equation}

Therefore, if this bound holds, the predicted risk is distributionally robust under uncertainty with a probability of at least $1 - \epsilon$.

By filtering PENN predictions through epistemic and aleatoric uncertainty assessments, this framework ensures that only predictions with sufficient confidence and bounded distributional risk are retained, enabling risk-aware and reliable decision-making under uncertainty.

\subsection{MPC Controller}\label{MPC}
In this section, we define the safe set of system states $\mathcal{S}$ as the zero-superlevel set of a continuously differentiable function $\psi: \mathcal{X} \rightarrow \mathbb{R}$:
\begin{equation}
    \mathcal{S} = \left\{\mathbf{x}_t \in \mathcal{X} \mid \psi(\mathbf{x}_t) \geq 0 \right\},
\end{equation}
where $\mathbf{x}_t = [x_t, y_t]^\top$ denotes the robot's position at time step $t$, safety is interpreted as sufficient clearance from static and dynamic obstacles. It can be achieved by ensuring the robot's distance to any obstacle remains above a predefined threshold. Accordingly, we define the following function to encode this constraint and construct the safe set $\mathcal{S}$:
\begin{equation}
\psi(\mathbf{x}_t) = (x_t - \mathbf{x}^o_{t})^2 + (y_t - \mathbf{y}^o_{t})^2 - \eta_{\text{safe}}^2,
\end{equation}
where $[\mathbf{x}^o_{t}, \mathbf{y}^o_{t}]^\top$ denotes the position of the obstacle at time step $t$, and $\eta_{\text{safe}}$ represents the minimum required safety distance.

To enforce forward invariance of the safe set $\mathcal{S}$, we introduce a discrete-time control barrier function (CBF) constraint \cite{cbf1, cbf2}:
\begin{equation}\label{cbf}
\Delta \psi(\mathbf{x}_t, \mathbf{u}_t) + \xi \psi(\mathbf{x}_t) \geq 0,
\end{equation}
where $\Delta \psi(\mathbf{x}_t, \mathbf{u}_t) := \psi(\mathbf{x}_{t+1}) - \psi(\mathbf{x}_t)$, and $\xi \in (0,1]$ is a adjustable rate parameter.

This condition ensures that $\psi$ satisfies the requirements of a discrete-time CBF, such that the safe set $\mathcal{S}$ remains invariant under the system's evolution. Moreover, Eq.~\eqref{cbf} guarantees that the minimum value of $\psi$ decreases no faster than an exponential rate of $1 - \xi$, thus preventing sudden violations of safety boundaries.

We formulate the navigation task as a nonlinear model predictive control (NMPC) \cite{nmpc} problem with safety constraints. The optimization problem at each decision step is given by:
\begin{subequations}
\begin{align}
    \min_{\{\mathbf{x}_t, \mathbf{u}_t\}} \quad & \|\mathbf{x}_N - \mathbf{x}_{\text{goal}}\|_{Q}^2 + \sum_{j=0}^{N-1} \|\mathbf{u}_t\|_{R}^2 \label{a}\\
    \text{s.t.} \quad & \mathbf{x}_{t+1} = \mathcal{O}(\mathbf{x}_t, \mathbf{u}_t) \label{b}\\
    & \mathbf{x}_{t=0} = \mathbf{x}_{\text{init}} \label{c}\\
    & \mathbf{x}_t \in \mathcal{X}, \ \mathbf{u}_t \in \mathcal{U} \label{d}\\
    & \Delta \psi(\mathbf{x}_t, \mathbf{u}_t) + \xi \psi(\mathbf{x}_t) \geq 0 \label{e}
\end{align}
\end{subequations}
  
Where $\|\mathbf{x}\|_H:= \sqrt{\frac{1}{2}\mathbf{x}^\top H \mathbf{x}}$ denotes the weighted norm, where $Q > 0$ and $R > 0$ are positive definite matrices representing the terminal cost and control effort penalty, respectively. 
The objective in Eq.~\eqref{a} minimizes a quadratic cost over a prediction horizon of $N$ steps. 
Eq.~\eqref{b} encodes the discrete-time system dynamics, modeled using omnidirectional-drive kinematics. 
Eq.~\eqref{d} ensures the state and control inputs remain within their feasible sets $\mathcal{X}$ and $\mathcal{U}$.
Eq.~\eqref{e} enforces CBF constraints for all obstacles, ensuring that safety margins are maintained throughout the horizon.
The above optimization problem is implemented in the CasADi framework~\cite{casadi} and solved in real-time using the IPOPT solver~\cite{ipopt}.

\subsection{Navigation}\label{Navigation}
\begin{figure}[htb]
\centering
    \includegraphics[width=1\columnwidth]{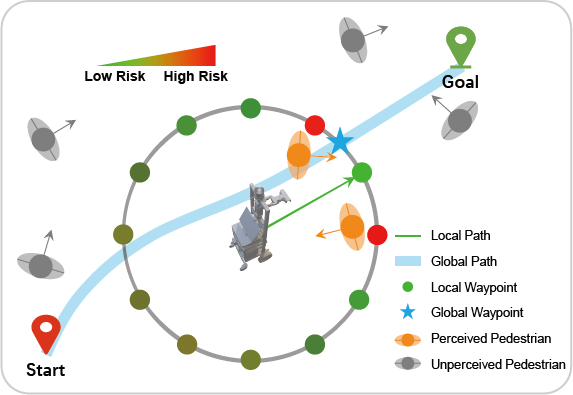}
\caption{Robot Crowd Navigation.}
\label{navigation}
\end{figure}

As illustrated in Fig.~\ref{navigation}, the diagram shows the robot’s navigation process in dynamic crowd environments. Orange human figures indicate humans detected within the robot’s sensor range, while gray ones represent undetected individuals. The blue trajectory shows the global path from start to goal, providing long-term navigational guidance. In contrast, the green dashed line illustrates the local path, which is continuously adjusted based on risk assessment. Blue stars denote global waypoints, whereas green dots represent candidate local waypoints sampled within the robot’s sensing field.
The PENN model evaluates each candidate waypoint to predict its associated risk. An uncertainty-aware filtering process follows, retaining only highly confident predictions and bounded uncertainty. The waypoint with the lowest estimated risk is selected as the next target for the robot. The robot continuously updates global and local paths throughout navigation, performs PENN risk inference, and applies real-time uncertainty filtering, enabling safe, adaptive, and socially aware navigation in dynamic crowd environments.

\section{Experiments}\label{experiments}
This section presents the experimental setup, evaluation metrics, and results from both simulation and real-world tests. We describe the implementation details (Sec.~\ref{implementation details}), environment settings (Sec.~\ref{environment settings}), and baselines (Sec.~\ref{baseline comparisons}). Next, we introduce the evaluation metrics used to quantify navigation performance and social awareness (Sec.~\ref{metric}). Finally, extensive simulation studies (Sec.~\ref{simulation}) and physical robot experiments (Sec.~\ref{robot trial}) validate the effectiveness of the proposed approach.

\subsection{Implementation Details}\label{implementation details}
We generate a dataset of 39653 samples offline for model training by collecting robot navigation data in simulated crowd environments using the HR-MPC algorithm. Each sample contains robot and human state information and corresponding risk assessment labels. To simulate real-world sensor noise and model uncertainty, we augment the dataset by adding 5\% independent and identically distributed Gaussian noise to the input features. The proposed PENN model consists of an ensemble of three neural networks, each comprising five fully connected layers with ReLU \cite{relu} activation functions.
We implement the training process offline using PyTorch on an AMD EPYC MILAN 7413 CPU and an NVIDIA RTX A6000 GPU. The PENN model is trained with the Adam optimizer \cite{adam} and an initial learning rate of 0.0001. To enhance training stability, we employ a learning rate scheduler that reduces the learning rate by a factor of 0.5 if no improvement is observed in validation loss for five consecutive epochs. An early stopping criterion also stops training if no improvement is detected over 10 consecutive epochs. The batch size is set to 256, and training runs for a maximum of 1000 epochs.
\subsection{Environment Settings}\label{environment settings}
We evaluate the proposed method in a widely used crowd simulation environment \cite{attnrnn}, which is commonly adopted for benchmark crowd navigation algorithms. While most existing studies focus solely on dynamic agents, we enhance the environment by incorporating static obstacles. In particular, a circle static obstacle with a radius of 1 m is placed at the center of the environment, as shown in Fig. \ref{compare}. The robot needs to navigate from the start to the goal, avoiding both moving crowd agents and the static obstacle. Crowd agents are simulated using the SF model, with a radius of 0.3 m and a maximum speed of 1.0 m/s. The robot is modeled as an omnidirectional agent with a radius of 0.3 m, a maximum speed of 1.0 m/s, and a sensing range of 5 m. The simulation environment runs with a discrete time step of 0.25 seconds. Any navigation episode exceeding 200 seconds is considered a failure.
\subsection{Baseline Comparisons}\label{baseline comparisons}
We compare our proposed method with representative crowd navigation approaches: 
\begin{itemize}
    \item Reactive-based Method: ORCA \cite{orca} and SF \cite{sf}, two reactive navigation approaches based on velocity obstacles and social forces, respectively, which perform purely reactive navigation by responding to nearby agents.
    \item Learning-based Method: AttnRNN \cite{attnrnn}, a socially aware RL approach that leverages attention mechanisms and recurrent neural networks to model human-robot interactions for crowd navigation.
    \item Heuristic-based Risk Assessment Method: HR-MPC, a model predictive control framework that relies on predefined heuristic rules for estimating navigation risk in dynamic human environments.
\end{itemize}
\subsection{Performance Metrics}\label{metric}
We quantify navigation performance using metrics summarized in Tab. \ref{navigation_metrics}. Specifically, we report metrics for the overall performance across all episodes and separately for episodes where the robot successfully reached the goal. Additionally, we propose two socially aware metrics: (1) Zone Entry Ratio, measuring the proportion of humans entering defined proxemic zones (intimate, personal, social, and public), and (2) Zone Time Ratio, evaluating the average proportion of time humans spend within each proxemic zone during robot navigation.
\subsubsection{Navigation Metrics}
We define $N_{\text{total}}$ as the total number of episodes and $N_{\text{succ}}$ as the number of successful episodes. For each episode, $e$, $T_e$, and $L_e$ represent the navigation time and path length, respectively. 
\begin{table}[htb]
    \centering
    \caption{Definitions of navigation performance metrics}
    \label{navigation_metrics}
    \renewcommand{\arraystretch}{1.35}
    \resizebox{\columnwidth}{!}{
    \begin{tabular}{ll}
    \toprule[1pt]
    \textbf{Metric} & \textbf{Definition} \\
    \hline
    Success Rate & $\displaystyle SR = \frac{N_\text{succ}}{N_\text{total}}$ \\[6pt]
    Average Time & $\displaystyle \bar{T} = \frac{1}{N_\text{total}}\sum_{e=1}^{N_\text{total}}T_e$ \\[6pt]
    Average Path Length & $\displaystyle \bar{L} = \frac{1}{N_\text{total}}\sum_{e=1}^{N_\text{total}}L_e$ \\[6pt]
    Average Time (Successful) & $\displaystyle \bar{T}_\text{succ} = \frac{1}{N_\text{succ}}\sum_{e=1}^{N_\text{succ}}T_e^\text{succ}$ \\[6pt]
    Average Path Length (Successful) & $\displaystyle \bar{L}_\text{succ} = \frac{1}{N_\text{succ}}\sum_{e=1}^{N_\text{succ}}L_e^\text{succ}$ \\
    \bottomrule [1pt]
    \end{tabular}}
\end{table}
\subsubsection{Socially aware Metrics}
\begin{itemize}
    \item Zone Entry Ratio: We define a set of distance zones as:
\begin{equation}
\mathcal{Z} = \{\text{Intimate}, \text{Personal}, \text{Social}, \text{Public}\}.
\end{equation}
For each zone \( z \in \mathcal{Z} \), we define the proportion of humans entering zone \( z \) as:
\begin{equation}
\mathcal{Z}_\text{space} = \frac{1}{\sum_{e=1}^{N_\text{total}} H_e} \sum_{e=1}^{N_\text{total}} \sum_{\tau=1}^{H_e} \mathbf{1}_z(h_\tau^e),
\end{equation}
where \( \mathbf{1}_z(h_\tau^e) = 1 \) if human \( h_\tau^e \) enters zone \( z \) at any time during episode \( e \), and 0 otherwise. $H_e$ denotes the number of humans in the $e$-th episode.

\item Zone Time Ratio: Let \( T_\tau^e \) denote the total number of time step for human \(h_\tau^e\), and \(T_{z,\tau}^{e}\) be the number of time step spent in zone \( z \). Then, the average proportion of time humans spend in each zone \( z \) is defined as:
\begin{equation}
\mathcal{Z}_\text{time} = \frac{1}{\sum_{e=1}^{N_\text{total}}H_e}\sum_{e=1}^{N_\text{total}}\sum_{\tau=1}^{H_e}\frac{T_{z,\tau}^{e}}{T_\tau^{e}}.
\end{equation}
\end{itemize}
\subsection{Simulation Results}\label{simulation}
\begin{table}[h]
\centering
\caption{Navigation Performance in Obstacle-Free Environment with Humans Aware of the Robot.}
\label{no-ob-t}
\resizebox{\columnwidth}{!}{%
\renewcommand{\arraystretch}{1.3}
\begin{tabular}{cccccc}
\toprule[1pt]
{Method}  & \textbf{$SR\uparrow$} & \textbf{$\bar{T} (s)\downarrow$} & \textbf{$\bar{L} (m)\downarrow$} & \textbf{$\bar{T}_\text{succ} (s)\downarrow$} & \textbf{$\bar{L}_{succ} (m)\downarrow$} \\ \hline
{ORCA}    & 72\%                  & 42.02                       & 26.61                              & 39.20                               & 26.68                                      \\
{SF}      & 71\%                  & 42.36                       & \textbf{25.15}                              & 39.54                               & \textbf{25.74}                                      \\
{AttnRNN} & 74\%                  & 35.90                       & 50.60                              & \textbf{31.20}                               & 49.67                                      \\
{HR-MPC}  & 93\%                  & \textbf{34.82}                       & 29.55                              & 34.59                               & 29.31                                      \\
{LR-MPC}  & \textbf{98\%}                  & 35.47                       & 30.72                              & 35.19                               & 30.53                                      \\ \bottomrule [1pt]
\end{tabular}%
}
\end{table}

\begin{table}[h]
\centering
\caption{Navigation Performance in Obstacle Environment with Humans Aware of the Robot.}
\label{ob-t}
\resizebox{\columnwidth}{!}{%
\renewcommand{\arraystretch}{1.3}
\begin{tabular}{cccccc}
\toprule[1pt]
{Method}  & \textbf{$SR\uparrow$} & \textbf{$\bar{T} (s)\downarrow$} & \textbf{$\bar{L} (m)\downarrow$} & \textbf{$\bar{T}_\text{succ} (s)\downarrow$} & \textbf{$\bar{L}_{succ} (m)\downarrow$} \\ \hline
{ORCA}    & 65\%                  & 42.68                       & 26.74                              & 39.13                               & 26.78                                      \\
{SF}      & 57\%                  & 44.76                       & \textbf{24.45}                              & 41.37                               & \textbf{25.98}                                      \\
{AttnRNN} & 29\%                  & \textbf{25.43}                       & 36.12                              & \textbf{33.24}                               & 50.33                                      \\
{HR-MPC}  & 89\%                  & 35.29                       & 29.12                              & 34.71                               & 29.24                                      \\
{LR-MPC}  & \textbf{93\%}                  & 37.33                       & 32.54                              & 36.44                               & 31.87                                     \\ \bottomrule [1pt]
\end{tabular}%
}
\end{table}

\begin{table}[h]
\centering
\caption{Navigation Performance in Obstacle-Free Environment with Humans Unaware of the Robot.}
\label{no-ob-f}
\resizebox{\columnwidth}{!}{%
\renewcommand{\arraystretch}{1.3}
\begin{tabular}{cccccc}
\toprule[1pt]
{Method}  & \textbf{$SR\uparrow$} & \textbf{$\bar{T} (s)\downarrow$} & \textbf{$\bar{L} (m)\downarrow$} & \textbf{$\bar{T}_\text{succ} (s)\downarrow$} & \textbf{$\bar{L}_{succ} (m)\downarrow$} \\ \hline
{ORCA}    & 50\%                  & 44.40                       & 30.20                              & 39.54                               & 30.14                                      \\
{SF}      & 57\%                  & 39.86                       & \textbf{24.22}                              & 40.04                               & \textbf{27.61}                                      \\
{AttnRNN} & 72\%                  & 34.08                       & 49.92                              & \textbf{31.27}                               & 50.48                                      \\
{HR-MPC}  & 74\%                  & \textbf{30.21}                       & 25.70                              & 34.03                               & 28.83                                      \\
{LR-MPC}  & \textbf{81\%}                  & 34.22                       & 29.57                              & 35.66                               & 31.05                                             \\ \bottomrule [1pt]
\end{tabular}%
}
\end{table}

\begin{table}[h]
\centering
\caption{Navigation Performance in Obstacle Environment with Humans Unaware of the Robot.}
\label{ob-f}
\resizebox{\columnwidth}{!}{%
\renewcommand{\arraystretch}{1.3}
\begin{tabular}{cccccc}
\toprule[1pt]
{Method}  & \textbf{$SR\uparrow$} & \textbf{$\bar{T} (s)\downarrow$} & \textbf{$\bar{L} (m)\downarrow$} & \textbf{$\bar{T}_\text{succ} (s)\downarrow$} & \textbf{$\bar{L}_{succ} (m)\downarrow$} \\ \hline
{ORCA}    & 46\%                  & 45.40                       & 28.44                              & 40.88                               & 30.03                                      \\
{SF}      & 48\%                  & 42.19                       & \textbf{24.94}                              & 41.38                               & \textbf{28.97}                                      \\
{AttnRNN} & 21\%                  & \textbf{22.62}                       & 33.81                              & \textbf{32.12}                               & 51.85                                      \\
{HR-MPC}  & 71\%                  & 30.59                       & 25.24                              & 34.46                               & 29.08                                      \\
{LR-MPC}  & \textbf{80\%}                  & 34.44                       & 29.89                              & 34.97                               & 30.79                                                       \\ \bottomrule [1pt]
\end{tabular}%
}
\end{table}

\begin{figure}[h]
\centering
    \includegraphics[width=1\columnwidth]{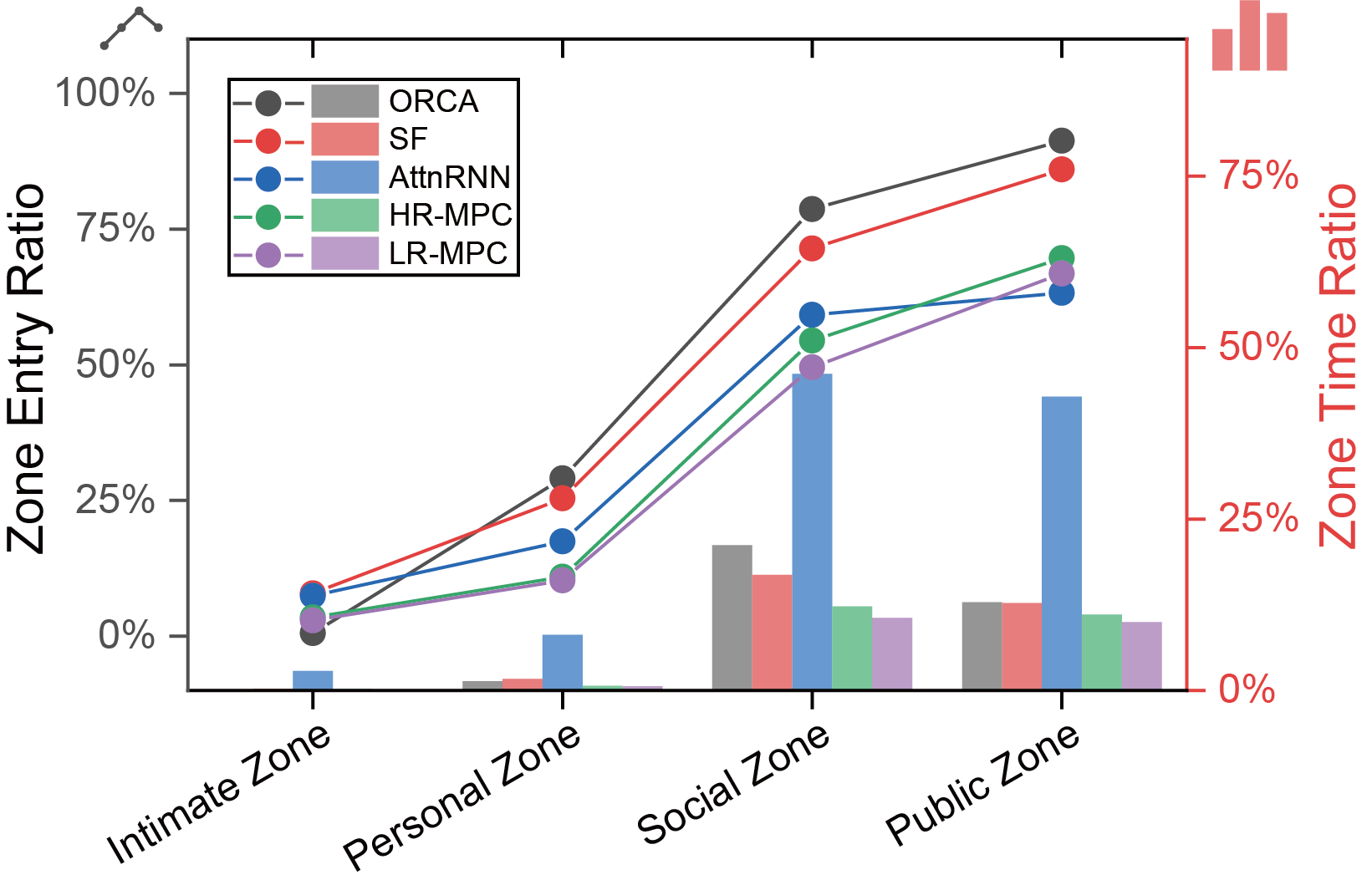}
\caption{Zone Entry and Time Ratios in Obstacle-Free Environment with Humans Unaware of the Robot.}
\label{no_ob_f}
\end{figure}

\begin{figure}[h]
\centering
    \includegraphics[width=1\columnwidth]{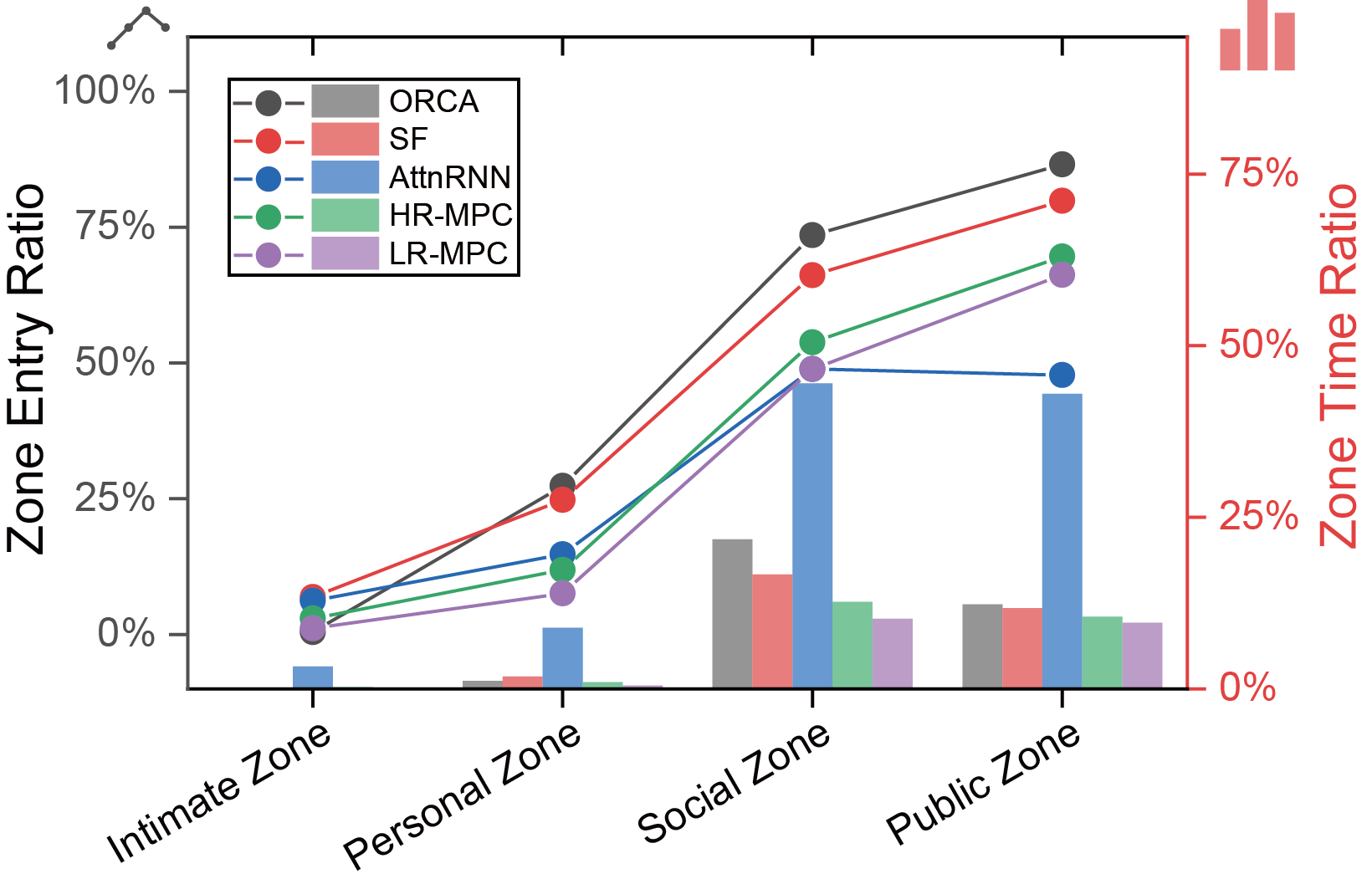}
\caption{Zone Entry and Time Ratios in Environment with Obstacles and Humans Unaware of the Robot.}
\label{ob_f}
\end{figure}

\begin{figure}[h]
\centering
    \includegraphics[width=1\columnwidth]{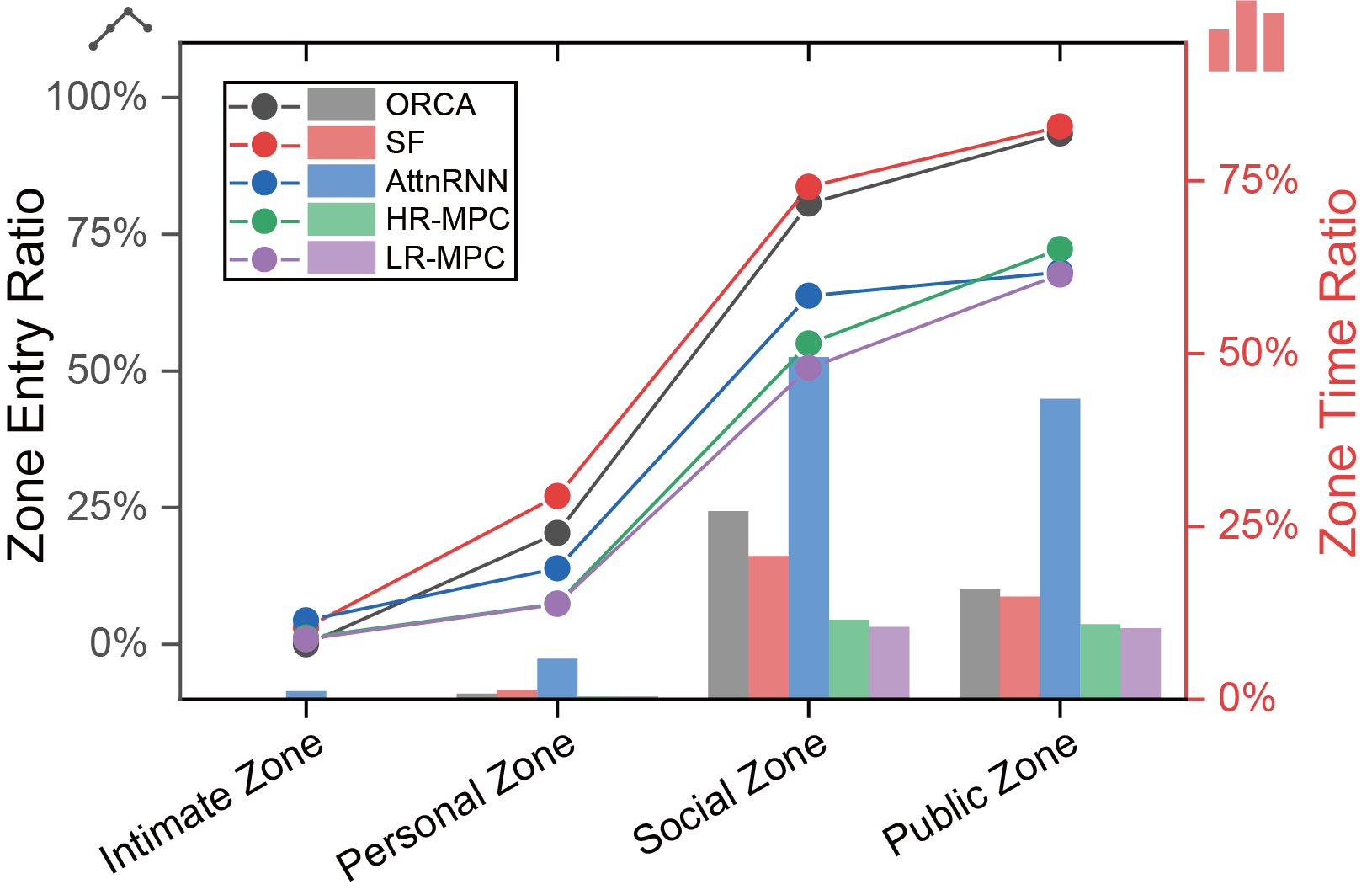}
\caption{Zone Entry and Time Ratios in Obstacle-Free Environment with Humans Aware of the Robot.}
\label{no_ob_t}
\end{figure}

\begin{figure}[h]
\centering
    \includegraphics[width=1\columnwidth]{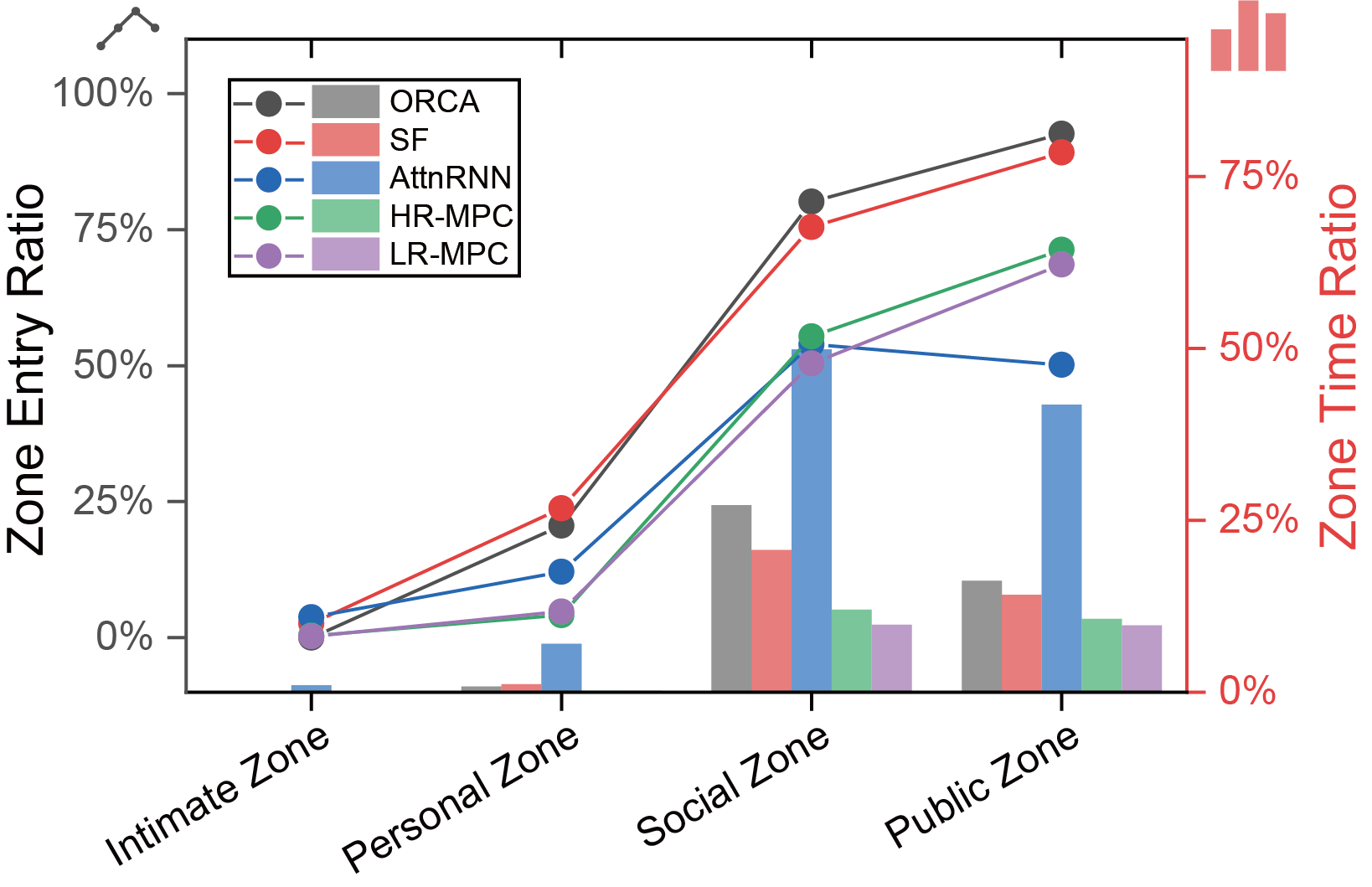}
\caption{Zone Entry and Time Ratios in Environment with Obstacles and Humans Aware of the Robot.}
\label{ob_t}
\end{figure}

We evaluate five navigation algorithms—ORCA, SF, AttnRNN, HR-MPC, and our proposed LR-MPC—across four simulation environments constructed by combining the presence of static obstacles and the crowd’s awareness of the robot. In the unaware setting, the crowd only avoids collisions with other humans without reacting to the robot—reflecting passive human-robot interaction. These combinations define four evaluation scenarios that reflect different complexities of human-robot interaction. In each simulation, the robot is tasked with navigating from the bottom-left to the top-right corner under the control of one of the five algorithms. The crowd size varies randomly between 5 and 30. For each human, both the initial and goal positions are randomly generated to ensure dynamic complexity. Each scenario is executed 100 times to ensure statistically reliable results. Performance metrics are collected across all episodes. Quantitative results on navigation metrics are reported in Tab. \ref{no-ob-t}-\ref{ob-f}, while socially aware metrics are visualized in Fig. \ref{no_ob_f}–\ref{ob_t}.

In Tables I–IV, LR-MPC consistently achieves the highest success rate across all scenarios, demonstrating strong robustness and adaptability. In contrast, AttnRNN exhibits significant performance degradation in complex settings, especially when static obstacles are present, and humans are unaware of the robot, where its success rate drops to as low as 21\%. HR-MPC performs more reliably than traditional baselines but falls short of LR-MPC in most metrics.

Regarding efficiency, LR-MPC maintains moderate average time and path length, indicating a balanced trade-off between navigation efficiency and safety. Although SF achieves the shortest paths in some settings, it suffers from lower success rates, suggesting potential collisions or failures due to its reactive nature. ORCA also shows longer navigation times and reduced reliability, particularly in uncooperative crowds.

Fig. \ref{no_ob_f}–Fig. \ref{ob_t} collectively present the zone entry ratios (black axis) and zone time ratios (red axis) across four proxemic zones—intimate, personal, social, and public—under varying environmental and human-awareness conditions. Five navigation strategies—ORCA, SF, AttnRNN, HR-MPC, and LR-MPC—are compared to evaluate their impact on human-robot spatial interaction.

Each figure corresponds to a specific scenario: Fig. \ref{no_ob_f} and Fig. \ref{no_ob_t} depict results in obstacle-free environments, with and without human awareness, respectively; Fig. \ref{ob_f} and Fig. \ref{ob_t} consider environments with static obstacles, again contrasting human awareness conditions. The results show that risk-aware methods (HR-MPC and LR-MPC) consistently reduce robot intrusions into closer human zones. Among them, LR-MPC demonstrates the highest level of social awareness, maintaining minimal entry and presence duration in intimate and personal zones.

Regardless of whether humans are aware of the robot's presence, our proposed methods maintain consistently low spatial intrusiveness, with LR-MPC achieving the best overall balance between safety and efficiency in dense crowd environments. LR-MPC demonstrates superior navigation performance, offering high success rates and socially aware trajectories, especially in dense and unstructured environments. These results validate the effectiveness of incorporating learned risk assessment and uncertainty filtering into the MPC framework.
\subsection{Robot Trial}\label{robot trial}
\begin{figure}[htb]
\centering
    \includegraphics[width=1.0\columnwidth]{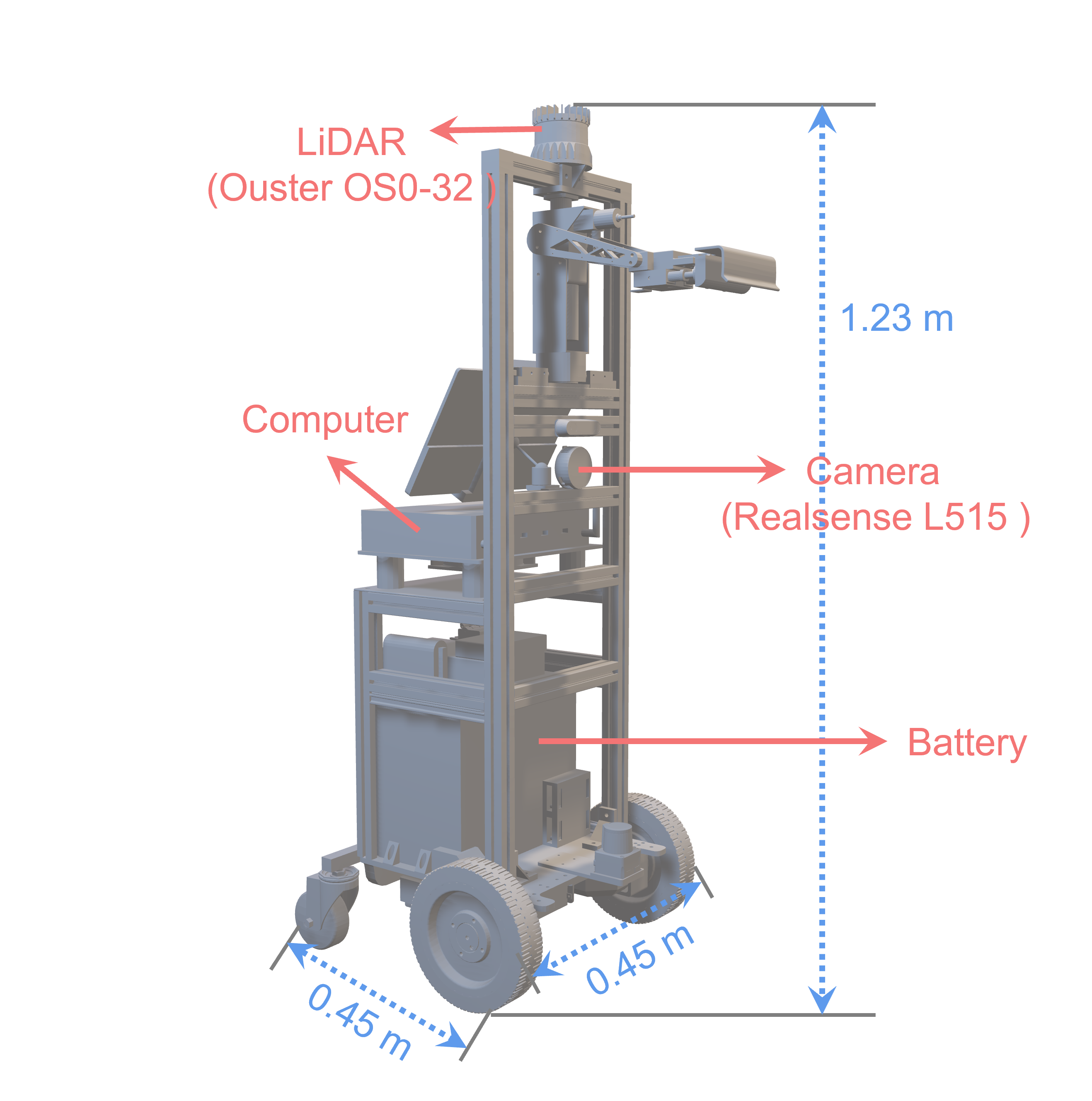}
\caption{Experimental platform for real-world experiment.}
\label{robot}
\end{figure}
\begin{figure*}[htb]
\centering
    \includegraphics[width=2\columnwidth]{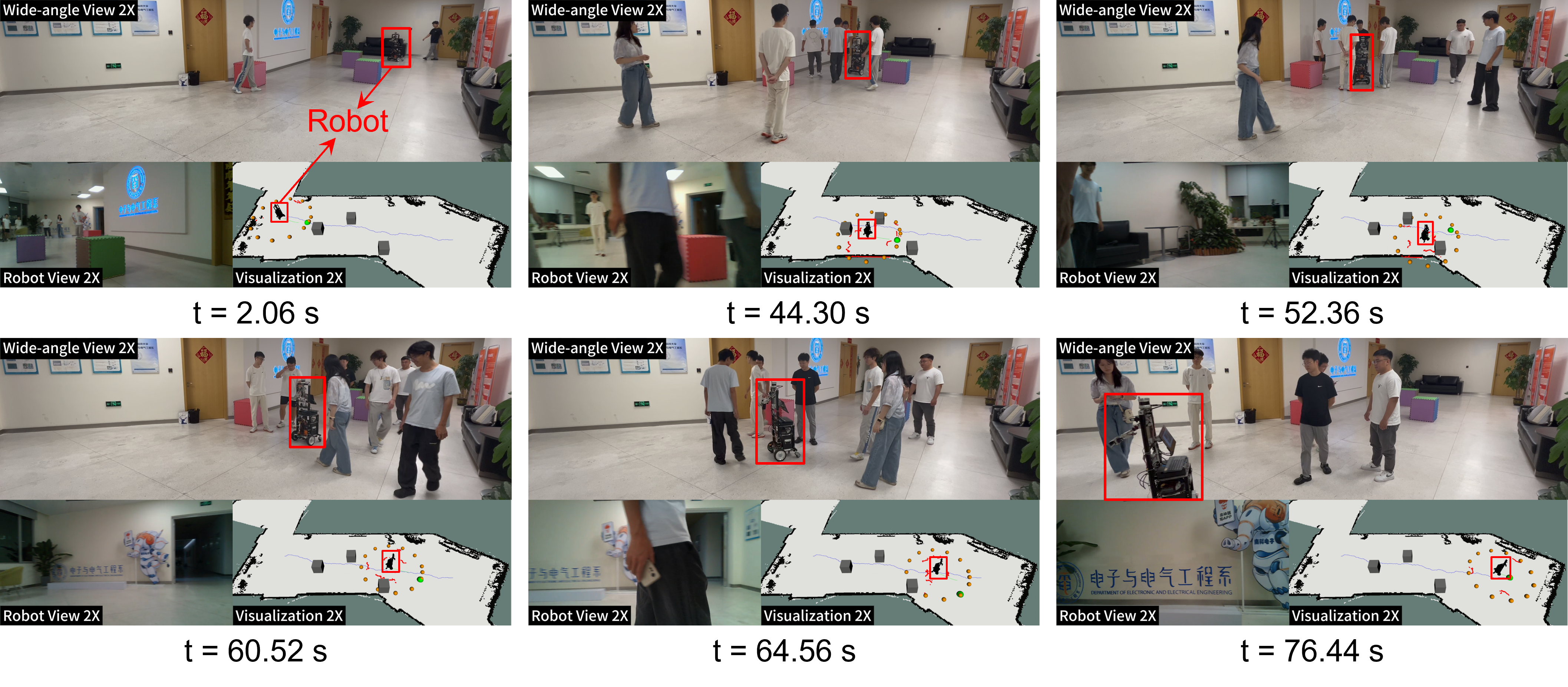}
\caption{Snapshots from real-world experiments using LR-MPC are presented at six representative time points: 2.06 s, 44.30 s, 52.36 s, 60.52 s, 64.56 s, and 76.44 s, demonstrating the robot navigating a dynamic crowded indoor environment. Each snapshot comprises a wide-angle view, a visualization view, and the robot's onboard perspective. A red bounding box in each image highlights the robot's position. In the visualization view, orange spheres indicate candidate evaluation waypoints, the green sphere indicates the local navigation waypoint, the blue line shows the global path generated by Multi-RRT, and the red dots correspond to LiDAR returns. Static obstacles are represented as cubes.}
\label{robot-trail}
\end{figure*}
The robot platform illustrated in Fig.~\ref{robot} has dimensions of 0.45 m × 0.45 m × 1.23 m. It features high-performance sensors such as an Ouster OS0-32 LiDAR and a Realsense L515 depth camera. Its onboard processing is handled by an Intel i9-12900F CPU paired with an NVIDIA RTX 3060 GPU.

In addition to the simulation experiments, we conducted real-world trials, as illustrated in Fig. \ref{robot-trail}, where the mobile robot employs the LR-MPC algorithm to navigate a dynamic, crowded indoor environment. The robot initiates from a predefined start location and autonomously navigates to a designated goal. Notably, seven moving humans actively interfered with the robot during the entire navigation process by continuously changing their positions. The LR-MPC algorithm effectively handled this by sampling candidate local waypoints (visualized as orange spheres in the visualization view) and evaluating their associated risk in real-time. At the same time, a global path (blue line in the visualization view) is generated by a Multi-RRT planner to guide the robot's long-term trajectory.

Continuous assessment and selection of optimal local waypoints enable the robot to keep a respectful distance from humans, ensuring socially aware navigation with minimal disruption. A complete experimental process video is available on the project website link\footnote{\href{https://sites.google.com/view/lr-mpc}{https://sites.google.com/view/lr-mpc}}.
\section{Discussion}
\begin{figure*}[htb]
\centering
    \includegraphics[width=2\columnwidth]{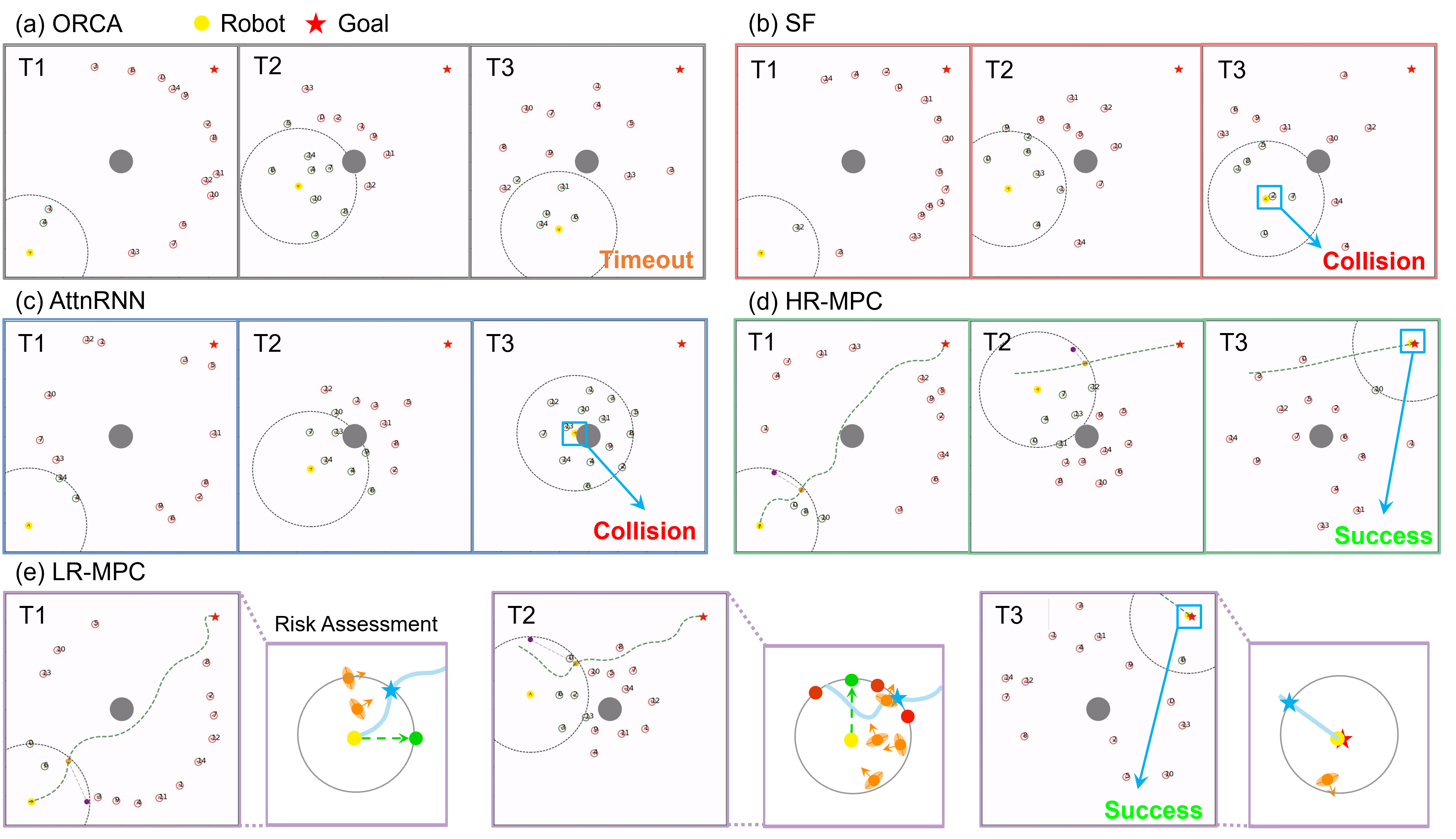}
\caption{Comparison of the navigation processes of five algorithms: (a) ORCA, (b) SF, (c) AttnRNN, (d) HR-MPC, and (e) LR-MPC. The yellow circle represents the robot, and the red pentagram denotes the goal. The dashed circle indicates the robot’s sensor range. The gray circle is the static obstacle. Dynamic humans are shown as numbered small circles: green if they are within the robot’s sensing range and red if outside.}
\label{compare}
\end{figure*}
In this section, we provide a detailed analysis of the performance improvements brought by LR-MPC, with a particular focus on its comparative advantages over existing navigation methods in dynamic crowd environments. As illustrated in Fig. \ref{compare}, the figure compares the navigation behaviors of five algorithms—ORCA, SF, AttnRNN, HR-MPC, and LR-MPC—across three representative time points (T1, T2, and T3). The experiments are conducted under conditions in which humans are unaware of the robot’s presence and do not actively avoid it, thus creating a purely reactive navigation scenario that tests the algorithm's robustness and adaptability.

The results reveal that ORCA and SF, as typical methods, heavily depend on reactive responses to human positions without risk assessment. Although ORCA manages to avoid obstacles initially, it lacks global foresight and ultimately becomes trapped in a local region, resulting in a timeout. Similarly, SF tends to local minima, producing overly conservative or unsafe avoidance paths. It leads to a collision in T3, highlighting its limitations in handling high-density dynamic environments.

AttnRNN leverages a graph-based attention mechanism to model agent interactions and exhibits aggressive avoidance behaviors. However, it lacks consideration for socially aware navigation principles. As observed at T3, the robot fails to effectively avoid static obstacles and dynamic crowds, leading to collisions. These outcomes highlight its limited robustness and inadequate perception of static structures in complex, crowded environments.

In contrast, both HR-MPC and LR-MPC can evaluate navigation risks proactively. From T1 to T3, both methods consistently generate feasible trajectories that successfully avoid densely crowded areas and complete the navigation task. However, HR-MPC depends on manually designed heuristic rules for risk assessment, which, while effective in structured scenarios, tend to be inflexible and struggle to adapt to unexpected environmental changes. LR-MPC adopts a data-driven strategy that enables online risk adaptation guided by epistemic and aleatoric uncertainty. It dynamically samples candidate waypoints during navigation, evaluates their risks, and selects the optimal local goal waypoint accordingly. At T2, the robot is seen actively steering away from high-interference zones, and at T3, it reaches the target efficiently. These results highlight LR-MPC's enhanced adaptability and ability to maintain socially aware behavior in dynamic, complex environments.

In summary, LR-MPC enables the robot to autonomously select the least risky local waypoint, achieving safe, efficient, and socially aware navigation in densely crowded environments.

\section{Conclusions and Future Work}
This article presents LR-MPC, an online uncertainty-driven risk adaptation algorithm for robot navigation in crowded environments. A data-driven model (PENN) predicts navigation risk, while a global planner (Multi-RRT) provides long-term guidance. Navigation risks are filtered to ensure robust decision-making using epistemic and aleatoric uncertainty, retaining only high-confidence predictions. The waypoint with the lowest estimated risk is selected as input to the MPC controller for real-time control generation. Experimental results demonstrate that LR-MPC enables adaptive risk-aware navigation in dynamic, crowded environments, achieving higher success rates and improved social awareness.

Future work will explore integrating large language models (LLMs) to enhance human-robot interaction. It will help robots better interpret human intent, express their state clearly, and reduce misunderstandings, improving cooperation and interaction in dense crowd environments.

\bibliographystyle{IEEEtran} 
\bibliography{refs} 

\begin{thebibliography}{10}
\providecommand{\url}[1]{#1}
\csname url@samestyle\endcsname
\providecommand{\newblock}{\relax}
\providecommand{\bibinfo}[2]{#2}
\providecommand{\BIBentrySTDinterwordspacing}{\spaceskip=0pt\relax}
\providecommand{\BIBentryALTinterwordstretchfactor}{4}
\providecommand{\BIBentryALTinterwordspacing}{\spaceskip=\fontdimen2\font plus
\BIBentryALTinterwordstretchfactor\fontdimen3\font minus
  \fontdimen4\font\relax}
\providecommand{\BIBforeignlanguage}[2]{{%
\expandafter\ifx\csname l@#1\endcsname\relax
\typeout{** WARNING: IEEEtran.bst: No hyphenation pattern has been}%
\typeout{** loaded for the language `#1'. Using the pattern for}%
\typeout{** the default language instead.}%
\else
\language=\csname l@#1\endcsname
\fi
#2}}
\providecommand{\BIBdecl}{\relax}
\BIBdecl

\bibitem{TMC1}
L.~Yu, Q.~Wang, Y.~Qiu, J.~Wang, X.~Zhang, and Z.~Han, ``Effective multi-agent
  communication under limited bandwidth,'' \emph{IEEE Transactions on Mobile
  Computing}, vol.~23, no.~7, pp. 7771--7784, 2024.

\bibitem{TMC2}
J.~Chen, J.~Cao, Z.~Cheng, and S.~Jiang, ``Towards efficient distributed
  collision avoidance for heterogeneous mobile robots,'' \emph{IEEE
  Transactions on Mobile Computing}, vol.~23, no.~5, pp. 3605--3619, 2024.

\bibitem{orca}
J.~Van Den~Berg, S.~J. Guy, M.~Lin, and D.~Manocha, ``Reciprocal n-body
  collision avoidance,'' in \emph{Robotics Research: The 14th International
  Symposium ISRR}.\hskip 1em plus 0.5em minus 0.4em\relax Springer, 2011, pp.
  3--19.

\bibitem{sf}
D.~Helbing and P.~Molnar, ``Social force model for pedestrian dynamics,''
  \emph{Physical review E}, vol.~51, no.~5, p. 4282, 1995.

\bibitem{TMC3}
Z.~Ning, H.~Ji, X.~Wang, E.~C.~H. Ngai, L.~Guo, and J.~Liu, ``Joint
  optimization of data acquisition and trajectory planning for uav-assisted
  wireless powered internet of things,'' \emph{IEEE Transactions on Mobile
  Computing}, vol.~24, no.~2, pp. 1016--1030, 2025.

\bibitem{shuijing0}
C.~Chen, Y.~Liu, S.~Kreiss, and A.~Alahi, ``Crowd-robot interaction:
  Crowd-aware robot navigation with attention-based deep reinforcement
  learning,'' in \emph{2019 international conference on robotics and automation
  (ICRA)}.\hskip 1em plus 0.5em minus 0.4em\relax IEEE, 2019, pp. 6015--6022.

\bibitem{shuijing1}
S.~Liu, P.~Chang, W.~Liang, N.~Chakraborty, and K.~Driggs-Campbell,
  ``Decentralized structural-rnn for robot crowd navigation with deep
  reinforcement learning,'' in \emph{2021 IEEE international conference on
  robotics and automation (ICRA)}.\hskip 1em plus 0.5em minus 0.4em\relax IEEE,
  2021, pp. 3517--3524.

\bibitem{attnrnn}
S.~Liu, P.~Chang, Z.~Huang, N.~Chakraborty, K.~Hong, W.~Liang, D.~L. McPherson,
  J.~Geng, and K.~Driggs-Campbell, ``Intention aware robot crowd navigation
  with attention-based interaction graph,'' in \emph{2023 IEEE international
  conference on robotics and automation (ICRA)}.\hskip 1em plus 0.5em minus
  0.4em\relax IEEE, 2023, pp. 12\,015--12\,021.

\bibitem{people_distance}
E.~T. Hall, R.~L. Birdwhistell, B.~Bock, P.~Bohannan, A.~R. Diebold~Jr,
  M.~Durbin, M.~S. Edmonson, J.~Fischer, D.~Hymes, S.~T. Kimball \emph{et~al.},
  ``Proxemics [and comments and replies],'' \emph{Current anthropology},
  vol.~9, no. 2/3, pp. 83--108, 1968.

\bibitem{ccr}
W.~Chi, H.~Kono, Y.~Tamura, A.~Yamashita, H.~Asama, and M.~Q.-H. Meng, ``A
  human-friendly robot navigation algorithm using the risk-rrt approach,'' in
  \emph{2016 IEEE International Conference on Real-time Computing and Robotics
  (RCAR)}.\hskip 1em plus 0.5em minus 0.4em\relax IEEE, 2016, pp. 227--232.

\bibitem{asymmetric_Gaussian}
W.~Chen, T.~Zhang, and Y.~Zou, ``Mobile robot path planning based on social
  interaction space in social environment,'' \emph{International Journal of
  Advanced Robotic Systems}, vol.~15, no.~3, p. 1729881418776183, 2018.

\bibitem{cai_gmm}
K.~Cai, W.~Chen, C.~Wang, S.~Song, and M.~Q.-H. Meng, ``Human-aware path
  planning with improved virtual doppler method in highly dynamic
  environments,'' \emph{IEEE Transactions on Automation Science and
  Engineering}, vol.~20, no.~2, pp. 1304--1321, 2022.

\bibitem{graph_optimization}
Z.~Jian, S.~Zhang, L.~Sun, W.~Zhan, N.~Zheng, and M.~Tomizuka, ``Long-term
  dynamic window approach for kinodynamic local planning in static and crowd
  environments,'' \emph{IEEE Robotics and Automation Letters}, vol.~8, no.~6,
  pp. 3294--3301, 2023.

\bibitem{penn}
T.~Kim, J.~Mun, J.~Seo, B.~Kim, and S.~Hong, ``Bridging active exploration and
  uncertainty-aware deployment using probabilistic ensemble neural network
  dynamics,'' \emph{arXiv preprint arXiv:2305.12240}, 2023.

\bibitem{penn2}
T.~Kim, R.~I. Kee, and D.~Panagou, ``Learning to refine input constrained
  control barrier functions via uncertainty-aware online parameter
  adaptation,'' \emph{arXiv preprint arXiv:2409.14616}, 2024.

\bibitem{uncertainty}
K.~Chua, R.~Calandra, R.~McAllister, and S.~Levine, ``Deep reinforcement
  learning in a handful of trials using probabilistic dynamics models,''
  \emph{Advances in neural information processing systems}, vol.~31, 2018.

\bibitem{dwa}
D.~Fox, W.~Burgard, and S.~Thrun, ``The dynamic window approach to collision
  avoidance,'' \emph{IEEE robotics \& automation magazine}, vol.~4, no.~1, pp.
  23--33, 2002.

\bibitem{rvo}
J.~Van~den Berg, M.~Lin, and D.~Manocha, ``Reciprocal velocity obstacles for
  real-time multi-agent navigation,'' in \emph{2008 IEEE international
  conference on robotics and automation}.\hskip 1em plus 0.5em minus
  0.4em\relax Ieee, 2008, pp. 1928--1935.

\bibitem{teb}
C.~R{\"o}smann, F.~Hoffmann, and T.~Bertram, ``Kinodynamic trajectory
  optimization and control for car-like robots,'' in \emph{2017 IEEE/RSJ
  International Conference on Intelligent Robots and Systems (IROS)}.\hskip 1em
  plus 0.5em minus 0.4em\relax IEEE, 2017, pp. 5681--5686.

\bibitem{Timed-ESDT}
D.~Zhu, T.~Zhou, J.~Lin, Y.~Fang, and M.~Q.-H. Meng, ``Online state-time
  trajectory planning using timed-esdf in highly dynamic environments,'' in
  \emph{2022 International Conference on Robotics and Automation (ICRA)}, 2022,
  pp. 3949--3955.

\bibitem{multi-risk-rrt}
Z.~Sun, B.~Lei, P.~Xie, F.~Liu, J.~Gao, Y.~Zhang, and J.~Wang,
  ``Multi-risk-rrt: An efficient motion planning algorithm for robotic
  autonomous luggage trolley collection at airports,'' \emph{IEEE Transactions
  on Intelligent Vehicles}, vol.~9, no.~2, pp. 3450--3463, 2024.

\bibitem{gmr-rrt}
J.~Wang, T.~Li, B.~Li, and M.~Q.-H. Meng, ``Gmr-rrt*: Sampling-based path
  planning using gaussian mixture regression,'' \emph{IEEE Transactions on
  Intelligent Vehicles}, vol.~7, no.~3, pp. 690--700, 2022.

\bibitem{namr-rrt}
Z.~Sun, B.~Xia, P.~Xie, X.~Li, and J.~Wang, ``Namr-rrt: Neural adaptive motion
  planning for mobile robots in dynamic environments,'' \emph{IEEE Transactions
  on Automation Science and Engineering}, 2025.

\bibitem{teleoperation}
T.~He, Z.~Luo, W.~Xiao, C.~Zhang, K.~Kitani, C.~Liu, and G.~Shi, ``Learning
  human-to-humanoid real-time whole-body teleoperation,'' in \emph{2024
  IEEE/RSJ International Conference on Intelligent Robots and Systems (IROS)},
  2024, pp. 8944--8951.

\bibitem{pedestrian_trajectory1}
Z.~Huang, A.~Hasan, K.~Shin, R.~Li, and K.~Driggs-Campbell, ``Long-term
  pedestrian trajectory prediction using mutable intention filter and warp
  lstm,'' \emph{IEEE Robotics and Automation Letters}, vol.~6, no.~2, pp.
  542--549, 2020.

\bibitem{pedestrian_trajectory2}
J.~Yang, Y.~Chen, S.~Du, B.~Chen, and J.~C. Principe, ``Ia-lstm:
  Interaction-aware lstm for pedestrian trajectory prediction,'' \emph{IEEE
  Transactions on Cybernetics}, vol.~54, no.~7, pp. 3904--3917, 2024.

\bibitem{TBD_dataset}
A.~Wang, D.~Sato, Y.~Corzo, S.~Simkin, A.~Biswas, and A.~Steinfeld, ``Tbd
  pedestrian data collection: Towards rich, portable, and large-scale natural
  pedestrian data,'' in \emph{2024 IEEE International Conference on Robotics
  and Automation (ICRA)}.\hskip 1em plus 0.5em minus 0.4em\relax IEEE, 2024,
  pp. 637--644.

\bibitem{drl-vo}
Z.~Xie and P.~Dames, ``Drl-vo: Learning to navigate through crowded dynamic
  scenes using velocity obstacles,'' \emph{IEEE Transactions on Robotics},
  vol.~39, no.~4, pp. 2700--2719, 2023.

\bibitem{social_principles}
A.~Francis, C.~P{\'e}rez-d’Arpino, C.~Li, F.~Xia, A.~Alahi, R.~Alami,
  A.~Bera, A.~Biswas, J.~Biswas, R.~Chandra \emph{et~al.}, ``Principles and
  guidelines for evaluating social robot navigation algorithms,'' \emph{ACM
  Transactions on Human-Robot Interaction}, vol.~14, no.~2, pp. 1--65, 2025.

\bibitem{socially_aware}
P.~T. Singamaneni, P.~Bachiller-Burgos, L.~J. Manso, A.~Garrell, A.~Sanfeliu,
  A.~Spalanzani, and R.~Alami, ``A survey on socially aware robot navigation:
  Taxonomy and future challenges,'' \emph{The International Journal of Robotics
  Research}, vol.~43, no.~10, pp. 1533--1572, 2024.

\bibitem{Sacson}
N.~Hirose, D.~Shah, A.~Sridhar, and S.~Levine, ``Sacson: Scalable autonomous
  control for social navigation,'' \emph{IEEE Robotics and Automation Letters},
  vol.~9, no.~1, pp. 49--56, 2023.

\bibitem{SocialGAIL}
B.~Ling, Y.~Lyu, D.~Li, G.~Gao, Y.~Shi, X.~Xu, and W.~Wu, ``Socialgail:
  Faithful crowd simulation for social robot navigation,'' in \emph{2024 IEEE
  International Conference on Robotics and Automation (ICRA)}, 2024, pp.
  16\,873--16\,880.

\bibitem{DR-MPC}
J.~R. Han, H.~Thomas, J.~Zhang, N.~Rhinehart, and T.~D. Barfoot, ``Dr-mpc: Deep
  residual model predictive control for real-world social navigation,''
  \emph{IEEE Robotics and Automation Letters}, 2025.

\bibitem{multi-rrt}
Z.~Sun, J.~Wang, and M.~Q.-H. Meng, ``Multi-tree guided efficient robot motion
  planning,'' \emph{Procedia Computer Science}, vol. 209, pp. 31--39, 2022.

\bibitem{renyi}
A.~R{\'e}nyi, ``On measures of entropy and information,'' in \emph{Proceedings
  of the fourth Berkeley symposium on mathematical statistics and probability,
  volume 1: contributions to the theory of statistics}, vol.~4.\hskip 1em plus
  0.5em minus 0.4em\relax University of California Press, 1961, pp. 547--562.

\bibitem{closed-form}
F.~Wang, T.~Syeda-Mahmood, B.~C. Vemuri, D.~Beymer, and A.~Rangarajan,
  ``Closed-form jensen-renyi divergence for mixture of gaussians and
  applications to group-wise shape registration,'' in \emph{Medical Image
  Computing and Computer-Assisted Intervention--MICCAI 2009: 12th International
  Conference, London, UK, September 20-24, 2009, Proceedings, Part I 12}.\hskip
  1em plus 0.5em minus 0.4em\relax Springer, 2009, pp. 648--655.

\bibitem{var}
T.~J. Linsmeier and N.~D. Pearson, ``Value at risk,'' \emph{Financial analysts
  journal}, vol.~56, no.~2, pp. 47--67, 2000.

\bibitem{cvar}
Y.~Chow, A.~Tamar, S.~Mannor, and M.~Pavone, ``Risk-sensitive and robust
  decision-making: a cvar optimization approach,'' \emph{Advances in neural
  information processing systems}, vol.~28, 2015.

\bibitem{cbf1}
J.~Zeng, B.~Zhang, and K.~Sreenath, ``Safety-critical model predictive control
  with discrete-time control barrier function,'' in \emph{2021 American Control
  Conference (ACC)}.\hskip 1em plus 0.5em minus 0.4em\relax IEEE, 2021, pp.
  3882--3889.

\bibitem{cbf2}
S.~He, J.~Zeng, B.~Zhang, and K.~Sreenath, ``Rule-based safety-critical control
  design using control barrier functions with application to autonomous lane
  change,'' in \emph{2021 American Control Conference (ACC)}.\hskip 1em plus
  0.5em minus 0.4em\relax IEEE, 2021, pp. 178--185.

\bibitem{nmpc}
A.~Xiao, H.~Luan, Z.~Zhao, Y.~Hong, J.~Zhao, W.~Chen, J.~Wang, and M.~Q.-H.
  Meng, ``Robotic autonomous trolley collection with progressive perception and
  nonlinear model predictive control,'' in \emph{2022 International Conference
  on Robotics and Automation (ICRA)}.\hskip 1em plus 0.5em minus 0.4em\relax
  IEEE, 2022, pp. 4480--4486.

\bibitem{casadi}
J.~A. Andersson, J.~Gillis, G.~Horn, J.~B. Rawlings, and M.~Diehl, ``Casadi: a
  software framework for nonlinear optimization and optimal control,''
  \emph{Mathematical Programming Computation}, vol.~11, pp. 1--36, 2019.

\bibitem{ipopt}
L.~T. Biegler and V.~M. Zavala, ``Large-scale nonlinear programming using
  ipopt: An integrating framework for enterprise-wide dynamic optimization,''
  \emph{Computers \& Chemical Engineering}, vol.~33, no.~3, pp. 575--582, 2009.

\bibitem{relu}
A.~F. Agarap, ``Deep learning using rectified linear units (relu),''
  \emph{arXiv preprint arXiv:1803.08375}, 2018.

\bibitem{adam}
D.~P. Kingma, ``Adam: A method for stochastic optimization,'' \emph{arXiv
  preprint arXiv:1412.6980}, 2014.

\end{thebibliography}
\begin{IEEEbiography}
[{\includegraphics[width=1in,height=1.25in,clip,keepaspectratio]{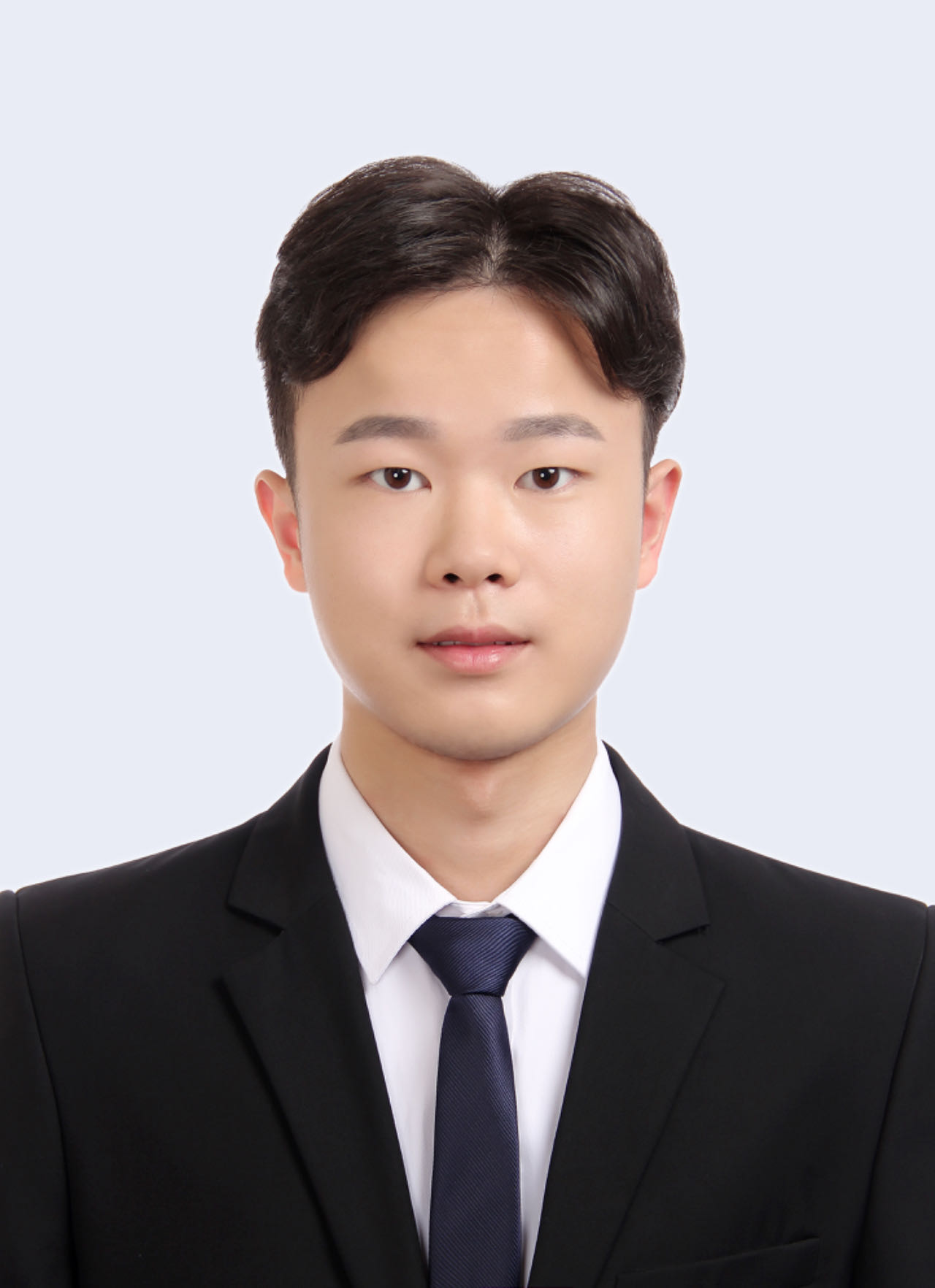}}] 
{Zhirui Sun} received the B.E. degree in information engineering from the Department of Electronic and Electrical Engineering, Southern University of Science and Technology, Shenzhen, China, in 2019. He is currently pursuing the Ph.D. degree with the Department of Electronic and Electrical Engineering, Southern University of Science and Technology, Shenzhen, China. His research interests include robot perception and motion planning.
\end{IEEEbiography}

\begin{IEEEbiography}
[{\includegraphics[width=1in,height=1.25in,clip,keepaspectratio]{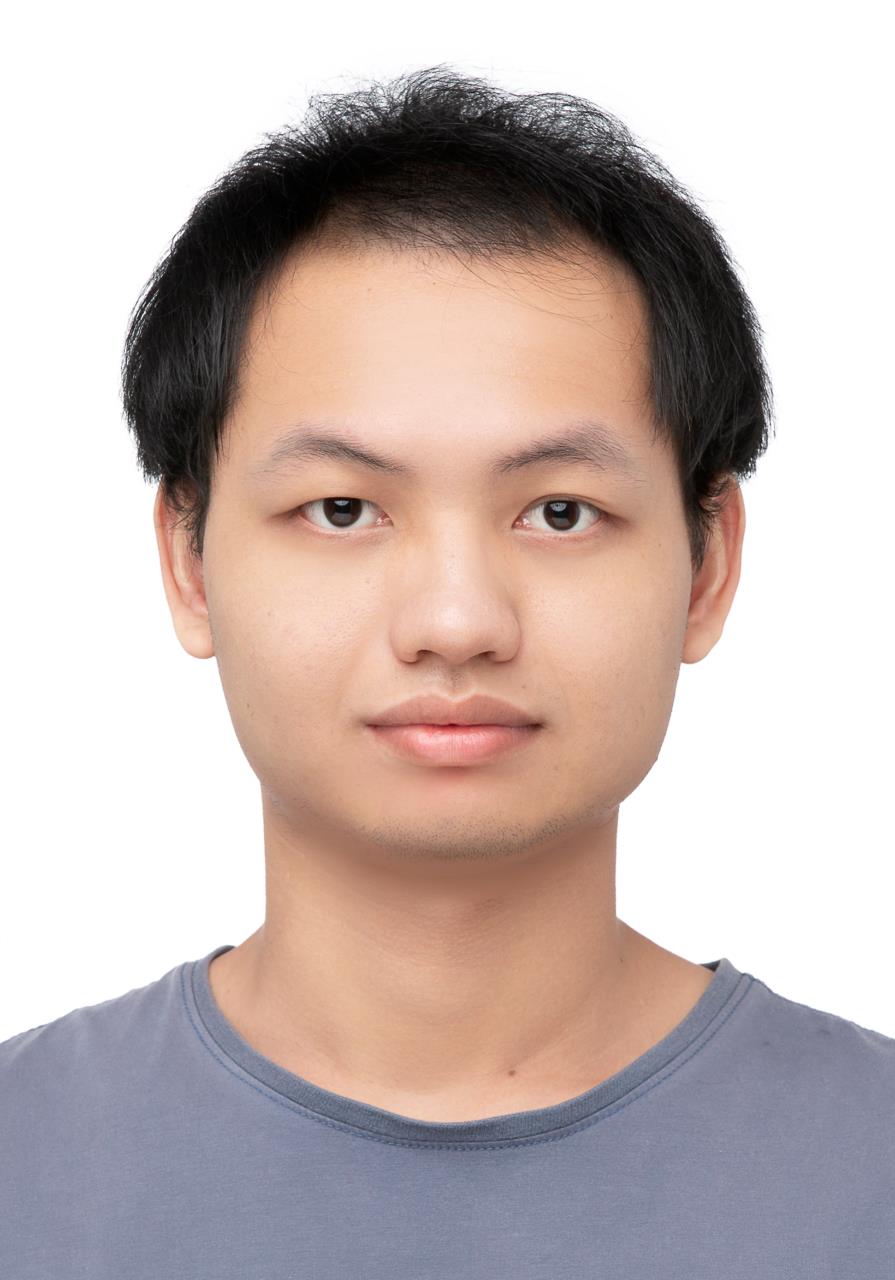}}] 
{Xingrong Diao} received the B.E degree in information engineering from the Department of Electronic and Electrical Engineering, Southern University of Science and Technology, Shenzhen, China, in 2023. He is currently pursuing the Master’s degree with the Department of Electronic and Electrical Engineering, Southern University of Science and Technology, Shenzhen, China. His research interests include robot manipulation and motion planning.
\end{IEEEbiography}

\begin{IEEEbiography}
[{\includegraphics[width=1in,height=1.25in,clip,keepaspectratio]{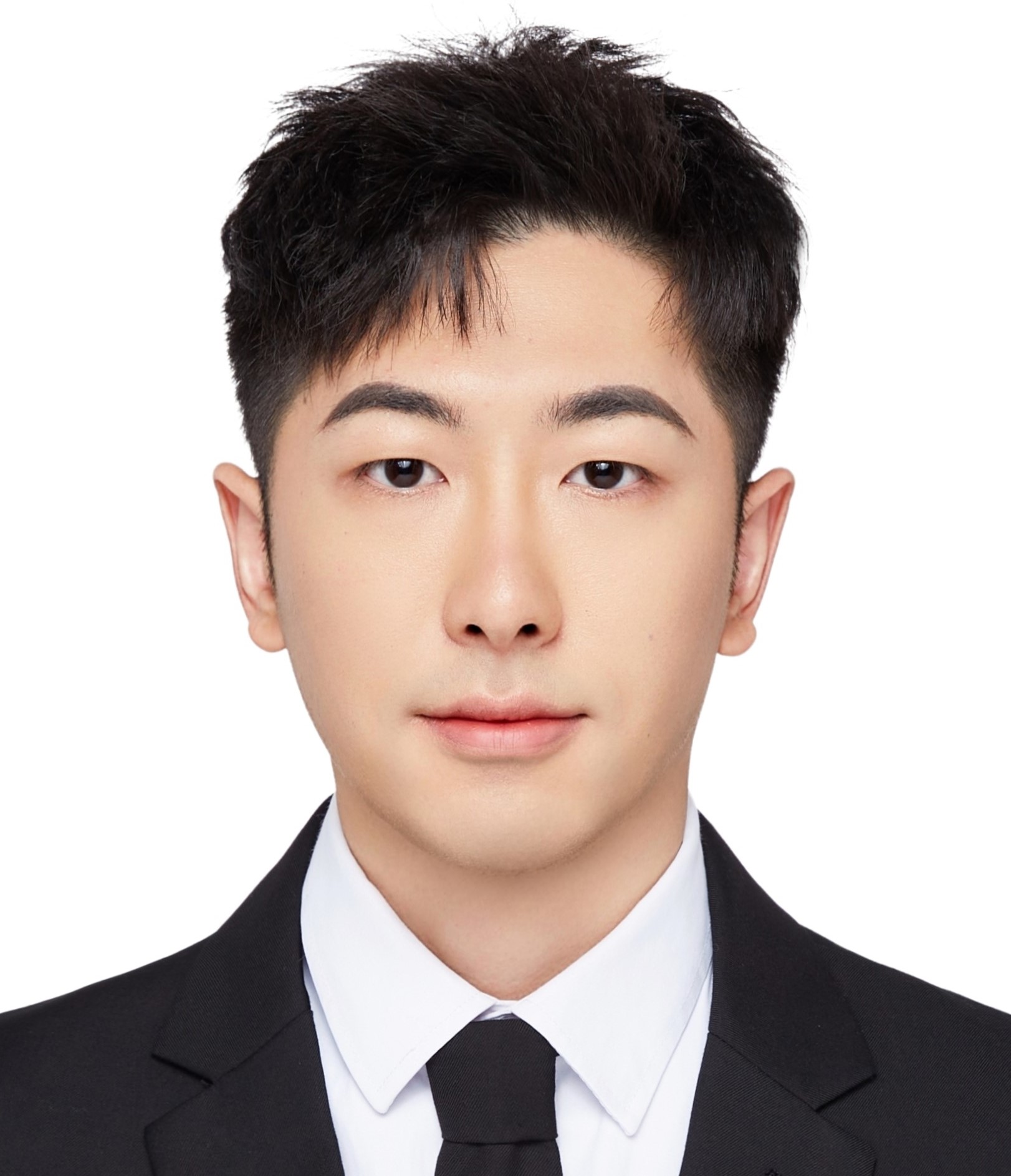}}] 
{Yao Wang} received the M.S. degree in mechanical engineering from the School of Mechanical and Electrical Engineering, Soochow University, Suzhou, China, in 2024. He is currently pursuing the Ph.D. degree with the Department of Electronic and Electrical Engineering, Southern University of Science and Technology, Shenzhen, China. His research field include mobile robot social navigation and visual language navigation.
\end{IEEEbiography}

\begin{IEEEbiography}
[{\includegraphics[width=1in,height=1.25in,clip,keepaspectratio]{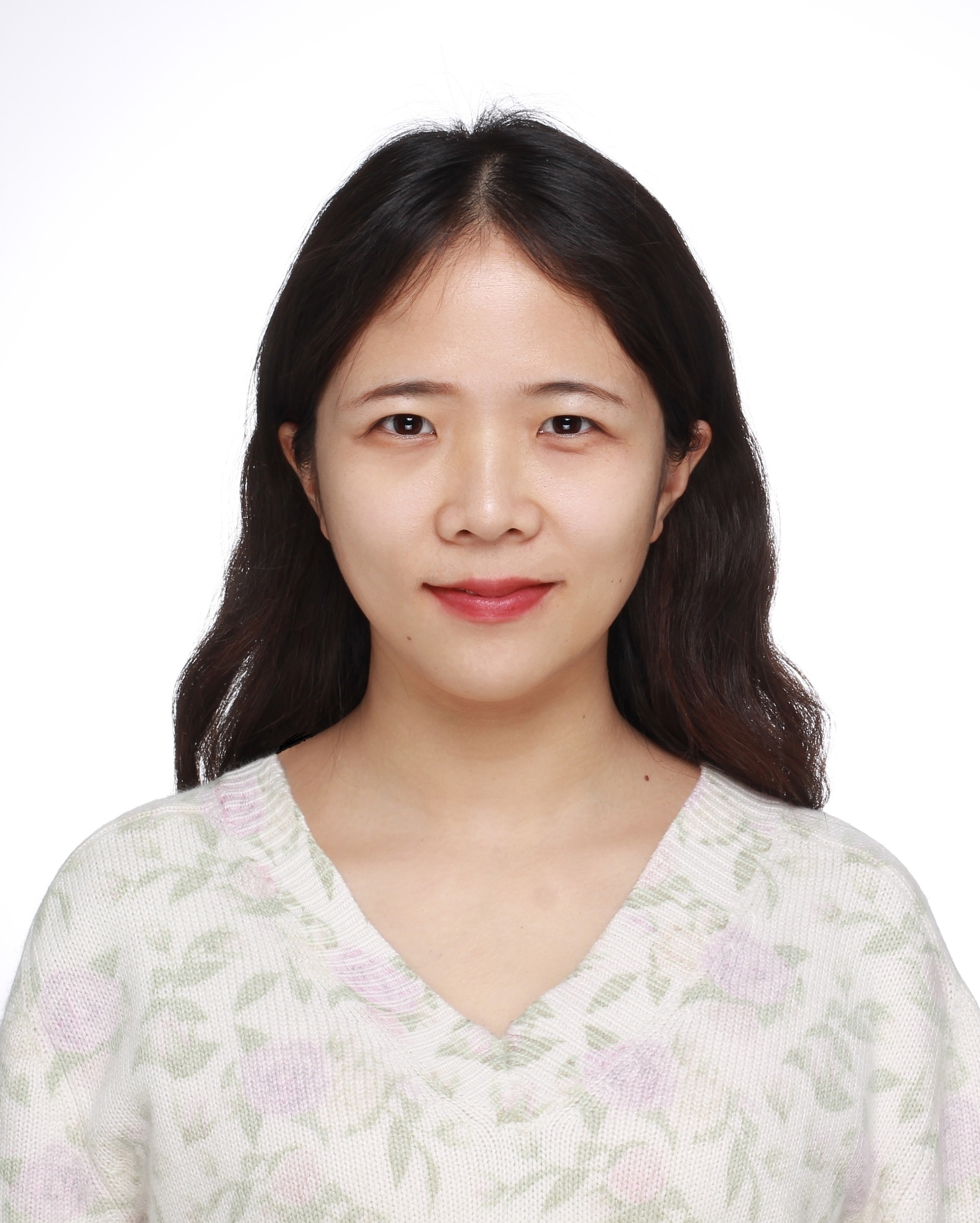}}] 
{Bi-Ke Zhu} received the B.E. degree in mechanical engineering from Dalian University of Technology, Dalian, China, in 2019, and Ph.D. degree in mechanical engineering from Shanghai Jiao Tong University, Shanghai, China, in 2024. 
She is currently a Postdoc researcher with the Department of Electronic and Electrical Engineering, Southern University of Science and Technology, Shenzhen, China. Her research interests include motion planning and intelligent control of robotic systems.

\end{IEEEbiography}

\begin{IEEEbiography}
[{\includegraphics[width=1in,height=1.25in,clip,keepaspectratio]{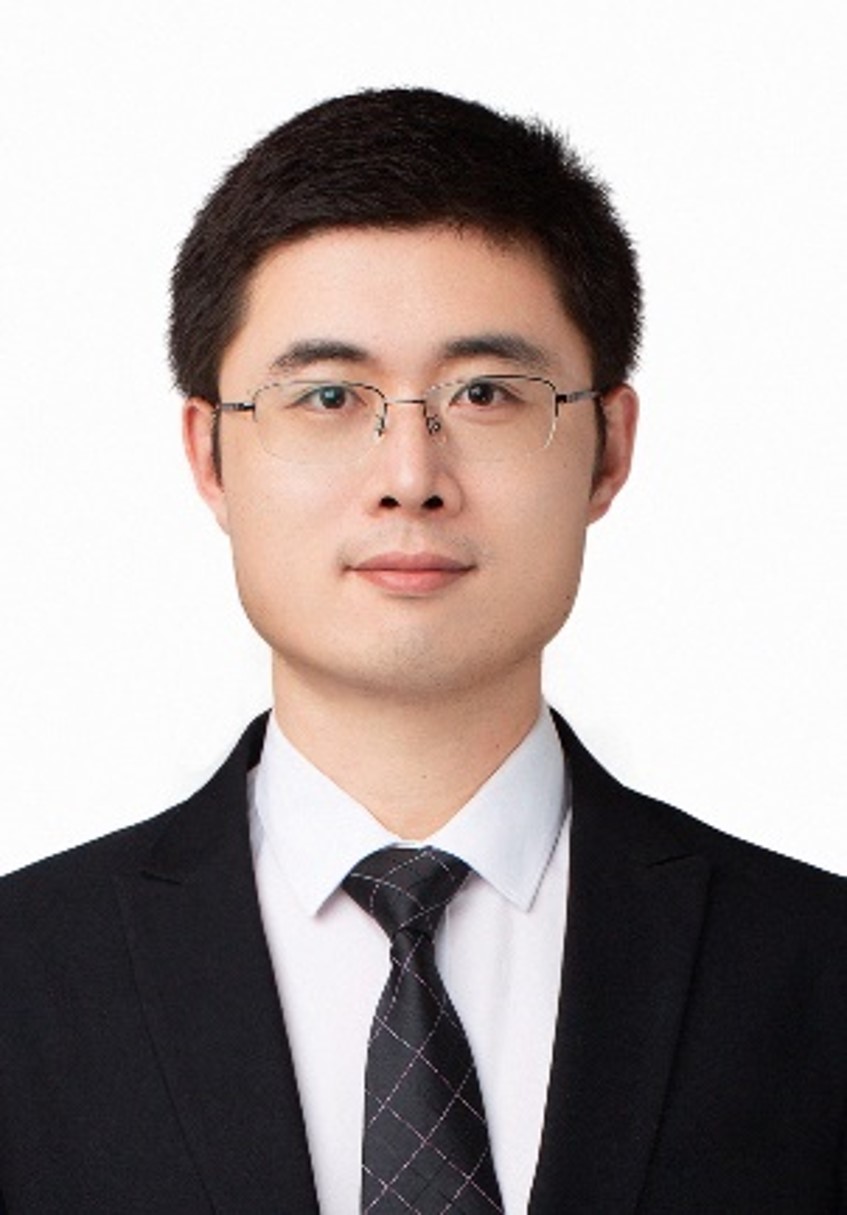}}] 
{Jiankun Wang} (Senior Member, IEEE) received the B.E. degree in automation from Shandong University, Jinan, China, in 2015, and the Ph.D. degree from the Department of Electronic Engineering, The Chinese University of Hong Kong, Hong Kong, in 2019.

He is currently an Assistant Professor with the Department of Electronic and Electrical Engineering, Southern University of Science and Technology, Shenzhen, China. His current research interests include motion planning and control, human–robot interaction, and machine learning in robotics.

Currently, he serves as the associate editor of IEEE Transactions on Automation Science and Engineering, IEEE Transactions on Intelligent Vehicles, IEEE Robotics and Automation Letters, International Journal of Robotics and Automation, and Biomimetic Intelligence and Robotics.
\end{IEEEbiography}

\end{document}